\documentclass[10pt,twocolumn,letterpaper]{article}

\usepackage[pagenumbers]{cvpr}      

\definecolor{cvprblue}{rgb}{0.21,0.49,0.74}
\usepackage[pagebackref,breaklinks,colorlinks,allcolors=cvprblue]{hyperref}

\usepackage{adjustbox}
\usepackage{array}

\newcolumntype{R}[2]{%
    >{\adjustbox{angle=#1,lap=\width-(#2)}\bgroup}%
    l%
    <{\egroup}%
}
\newcommand*\rot{\multicolumn{1}{R{80}{1em}}}

\usepackage{graphicx}
\usepackage{amsmath}
\usepackage{amssymb}
\usepackage{booktabs}

\usepackage{float}
\usepackage{tikz}
\usepackage{wrapfig}
\usepackage{colortbl}
\usepackage{pifont}

\def\our{GaMeS}
\def\mesh{Triangle Soup}

\def\R{\mathbb{R}}

\def\G{\mathcal{G}}
\def\M{\mathcal{M}}
\def\N{\mathcal{N}}
\def\m{\mathrm{m}}
\def\x{{\bf x}}
\def\m{{\bf m}}
\def\n{{\bf n}}
\def\F{\mathcal{F}}

\def\paperID{7223} 
\def\confName{CVPR}
\def\confYear{2025}

\title{\our{}: Mesh-Based Adapting and Modification of Gaussian Splatting}

\author{Joanna Waczyńska\thanks{Equal contribution}\\
  Faculty of Mathematics and Computer Science\\ 
  Jagiellonian University\\
      Doctoral School of Exact and Natural Sciences\\    
\and
Piotr Borycki*\\
  Faculty of Mathematics and Computer Science\\ 
  Jagiellonian University\\
\and
Sławomir Tadeja\\
  Department of Engineering\\
  University of Cambridge\\ 
\and
Jacek Tabor\\
  Faculty of Mathematics and Computer Science\\ 
  Jagiellonian University\\
\and
Przemysław Spurek\\
  Faculty of Mathematics and Computer Science\\ 
  Jagiellonian University\\
}

\begin{document}
\maketitle
\begin{abstract}
Gaussian Splatting (GS) is a novel, state-of-the-art technique for rendering points in a 3D scene by approximating their contribution to image pixels through Gaussian distributions, warranting fast training and real-time rendering. The main drawback of GS is the absence of a well-defined approach for its conditioning due to the necessity of conditioning several hundred thousand Gaussian components. To solve this, we introduce the \textbf{Ga}ussian \textbf{Me}sh \textbf{S}platting (\our{}) model, which allows modification of Gaussian components in a similar way as meshes. We parameterize each Gaussian component by the vertices of the mesh face. Furthermore, our model needs mesh initialization on input or estimated mesh during training. We also define Gaussian splats solely based on their location on the mesh, allowing for automatic adjustments in position, scale, and rotation during animation. As a result, we obtain a real-time rendering of editable GS.
\end{abstract}

\section{Introduction}

\begin{figure}[ht]
\vspace{1cm}
 	\centering
\includegraphics[width=0.45\textwidth, trim=10 10 10 10, clip]{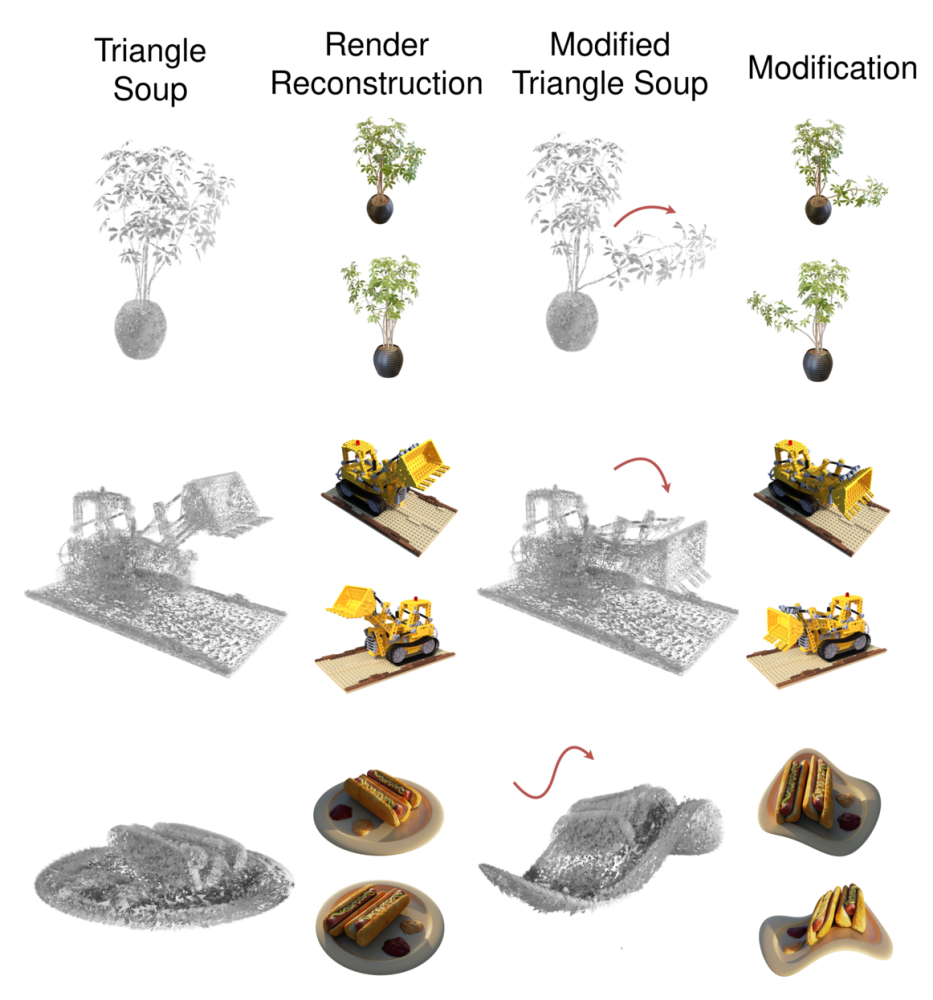}

 \caption{\our{} produce a hybrid of Gaussian Splatting (GS) and mesh representations. Therefore, \our{} allows real-time modification and adaptation of GS. }
 \vspace{0.5cm}
\label{fig:nerfsyntheticslego} 
\end{figure}

\begin{figure*}[h]
\includegraphics[width=0.95\textwidth]{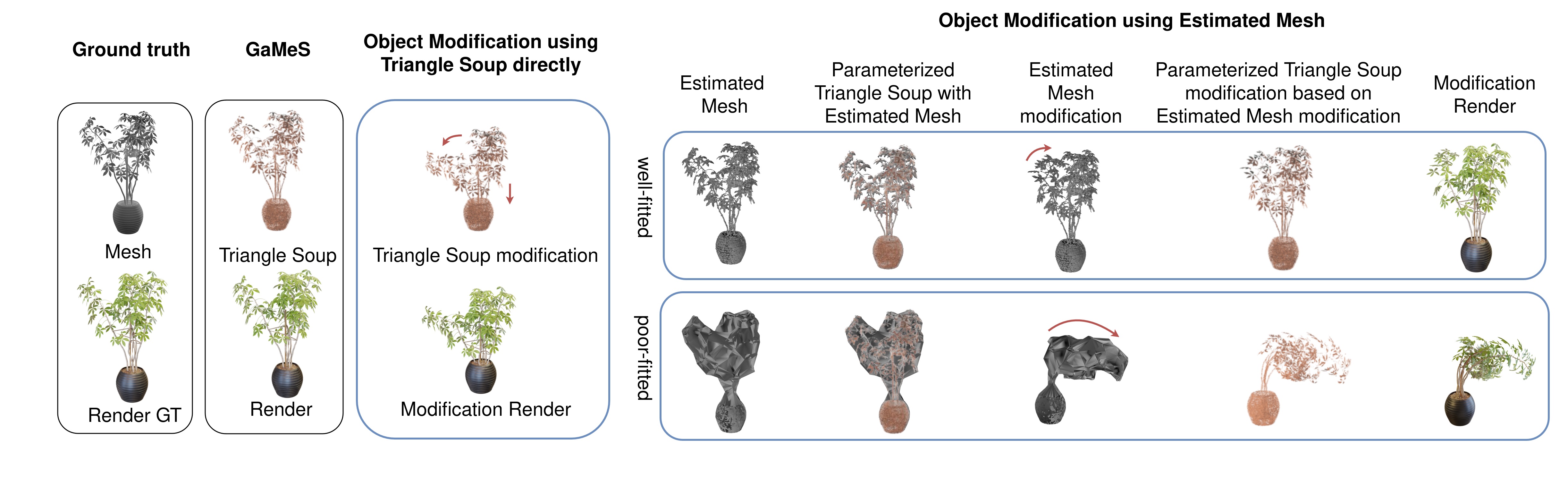}
\vspace{-0.5cm}
\caption{
In the absence of a pre-existing input mesh, we parameterize Gaussians to create a structure referred to as \mesh{}, which facilitates easy modification of the object. The Triangle Soup can be edited directly or modified using parametrization derived from estimated meshes, which do not need to be perfectly fitted. As a result, mesh-based deformation is applied to edit the object.}
\label{fig:edit} 
\end{figure*}

Recently, we have observed the emergence of several promising methods for rendering unseen views of 3D objects and scenes using neural networks. For instance, Neural Radiance Fields (NeRFs) \cite{mildenhall2020nerf} have rapidly grown in popularity within the computer vision and graphics communities~\cite{Gao2022NeRFNR} as they enable the creation of high-quality renders. Despite this interest and growing body of related research, the long training and inference time remains an unsolved challenge for NeRFs.

In contrast, latterly introduced Gaussian Splatting (GS)~\cite{kerbl20233d} offers swift training and real-time rendering capabilities. What is unique to this method is that it represents 3D objects using Gaussian distributions (i.e. Gaussians). Hence, it does not rely on any neural network. Consequently, Gaussians can be employed in a manner akin to manipulating 3D point clouds or meshes, allowing for actions like resizing or repositioning in 3D space. Nonetheless, practical challenges may arise when altering Gaussian positions, particularly in accurately tracking changes in the shape of Gaussian components, such as the ellipses. Moreover, scaling Gaussian components proves challenging when the object undergoes resizing, which is not an issue for classical meshes, as their triangle faces can be readily updated when vertex positions are adjusted.


\begin{figure}[h]
 Render Reconstruction \hspace{1cm}\ Modifications\\
 	\centering
 \includegraphics[width=0.49\textwidth, trim=0 0 0 0, clip]{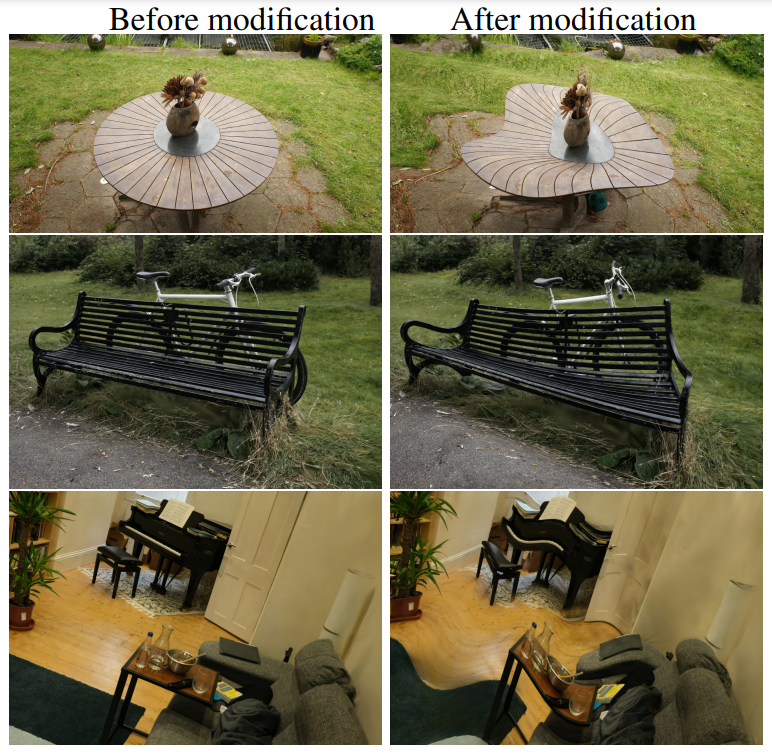} 
\caption{\our{} can be effectively trained on large scenes to allow their modifications while preserving high-quality renders.
}
\vspace{-0.3cm}
\label{fig:scanemod} 
\end{figure}

The above constraints may be resolved by constructing Gaussian directly on the mesh, as shown by SuGaR~\cite{guedon2023sugar}. Unlike SuGaR, we do not have two expensive stages dedicated to producing high-quality mesh. Instead, our \textit{Gaussian Mesh Splatting} (\our{}) is designed to accommodate two distinct input scenarios. First, \textit{mesh-initialized input}, when a mesh is provided as input, it may either be well-aligned (our baseline model) or poorly fitted, resembling the general form of the target object. For mismatched meshes, our one-stage training approach gradually refines the mesh alignment with the object during one-stage training. Second, \textit{mesh-free input} when no mesh is available, \our{} directly parametrizes flat Gaussians by constructing a representation inspired by triangular meshes. This results in a set of unconnected triangles called \mesh{}\footnote{\mesh{} can be called a pseudo-mesh because it imitates the mesh of a 3D object but lacks the explicit connectivity information that defines a mesh.}, which serves as a mesh-like structure.  In such a strategy, our \mesh{} is not dedicated to describing the object's surface, but we can treat \mesh{} analogously to a conventional mesh, allowing for flexible edits and object modifications (see Fig. \ref{fig:nerfsyntheticslego}). Additionally, the \our{} can be further parameterized by estimating an approximate mesh (see Fig. \ref{fig:edit}), providing a versatile foundation for structural adjustments and fine-grained control. The mesh-like approach can also be applied to model large scenes (see Fig.~\ref{fig:scanemod}).

Using \our{}, we can obtain comparable to state-of-the-art high-quality outcomes that can be attained for static scenes akin to the GS method. In summary, this work makes the following contributions:



\begin{itemize}
    \item We introduce a hybrid representation for 3D objects, seamlessly integrating mesh and GS. 
    \item We introduce a \mesh{} structure as a parametrization of flat GS for representing 3D objects, enabling efficient object editing without the need for costly meshing stages.
    \item Our method relies only upon essential vector operation. Consequently, we can render scenes at a similar pace to their static counterpart. 
\end{itemize}


\section{Related Works}


Point-based Gaussian representations have found a large number of potential application scenarios, including substituting point-cloud data \cite{eckart2016accelerated}, molecular structures modeling~\cite{blinn1982generalization}, or shape reconstruction \cite{keselman2023flexible}. These representations can also be utilized in applications involving shadow \cite{nulkar2001splatting} and cloud rendering \cite{man2006generating}. In addition, a new technique of 3D-GS was recently introduced \cite{kerbl20233d}. This method couples splatting methods with point-based rendering for real-time rendering speed. The rendering quality of GS is comparable to that of Mip-NeRF \cite{barron2021mip}, one of the best multilayer perceptron-based renderers.

\begin{figure}
\centering
\includegraphics[width=0.48\textwidth, trim = 0 0 0 200, clip]{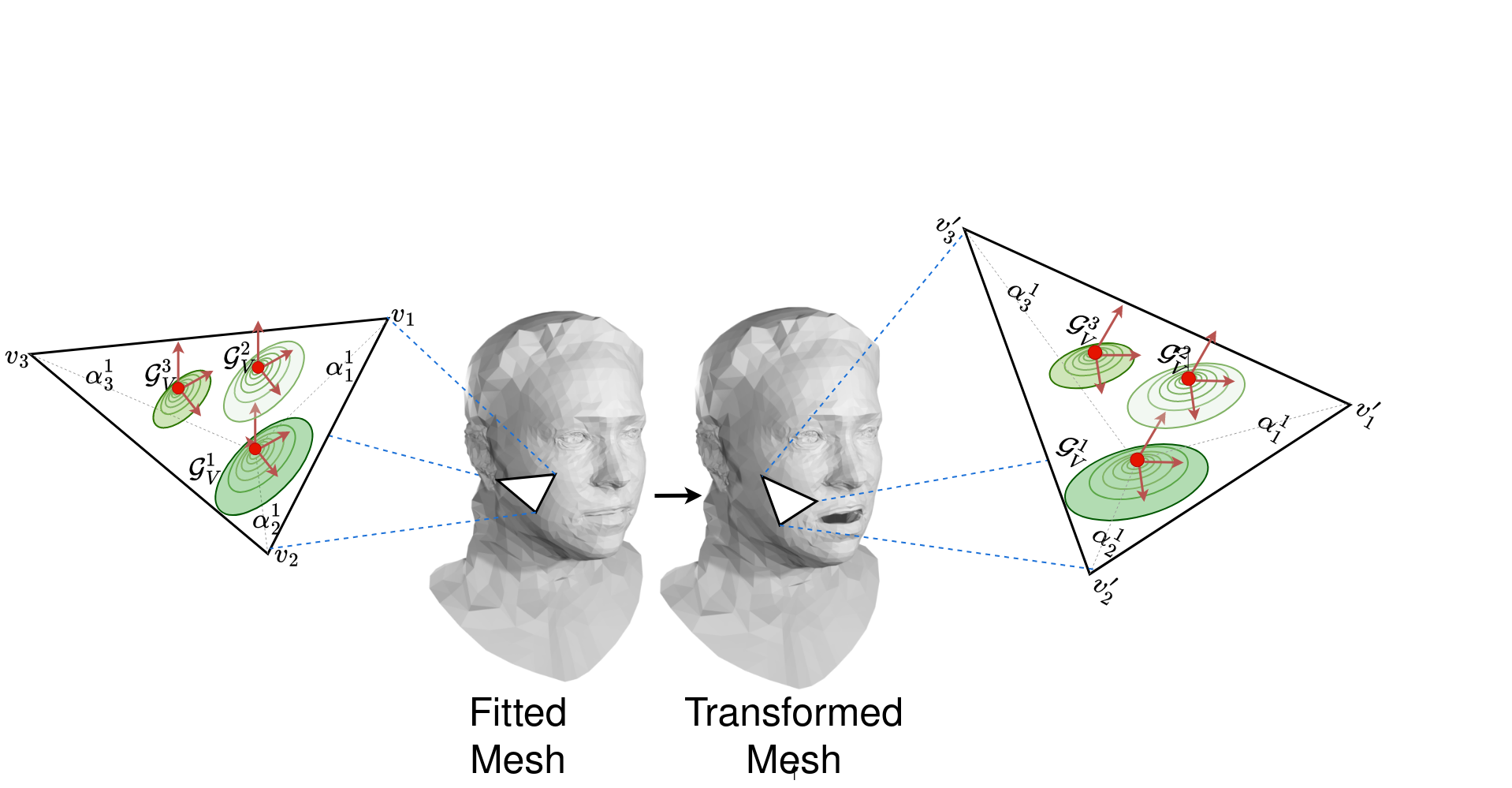}
\caption{Visualization of affine transformation of Gaussian components when we modify mesh. Mesh vertices parameterize the mean and covariance matrix of Gaussian. Therefore, such parameters update when we change the mesh.}
\label{fig:affinie} 
\end{figure}

The GS surpasses NeRF in terms of both training and inference speed, distinguishing itself by not depending on a neural network for operation. Instead, GS stores essential information within its 3D Gaussian components, a feature that renders it well-suited for dynamic scene modeling \cite{wu20234d}. Additionally, integrating GS with a dedicated 3D computer graphics engine is a straightforward process \cite{kerbl20233d}. However, conditioning GS is a challenging task due to the large number of Gaussian components typically involved, which can reach hundreds of thousands.

GaussianAvatars~\cite{qian2023gaussianavatars} utilizes the local coordinate system to generate Gaussians corresponding to the mesh's faces. This approach is designed explicitly for avatars and assumes the availability of a realistic (external) model for mesh fitting. However, training both the mesh and GS simultaneously is not feasible. While these solutions offer certain advantages, they do not directly combine Gaussian components with the mesh. As a result, automatic adaptation of the Gaussian parameters to changing mesh is impossible in these approaches.


The more advanced method is proposed by SuGaR~\cite{guedon2023sugar}, which introduces a regularization term in the GS cost function to encourage alignment between the Gaussians and the scene surface. SuGaR achieves this by using signed distance functions (SDF) and minimizing the difference between it and computed Gaussian values. Furthermore, the authors propose a method for extracting mesh from the GS model and introduce an optional refinement strategy that binds Gaussians to the mesh surface and jointly optimizes these Gaussians and the mesh through GS rendering.

In contrast, our work shows we can produce an editable GS model without expensive pre-processing stages like those used in SuGaR.


\section{\our{} : Mesh-Based Gaussian Splatting}


Here, we provide the details of the \our{} model, commencing with the fundamental aspects of vanilla GS. Subsequently, we elucidate how we parameterize Gaussian distributions on the mesh faces. Then, we investigate extracting a mesh face from a single Gaussian component used to re-parametrize Gaussian components to make them editable. Finally, we introduce our novel \our{} approach.

\begin{figure}
\centering
 \includegraphics[width=0.4\textwidth, trim=0 80 0 200, clip]{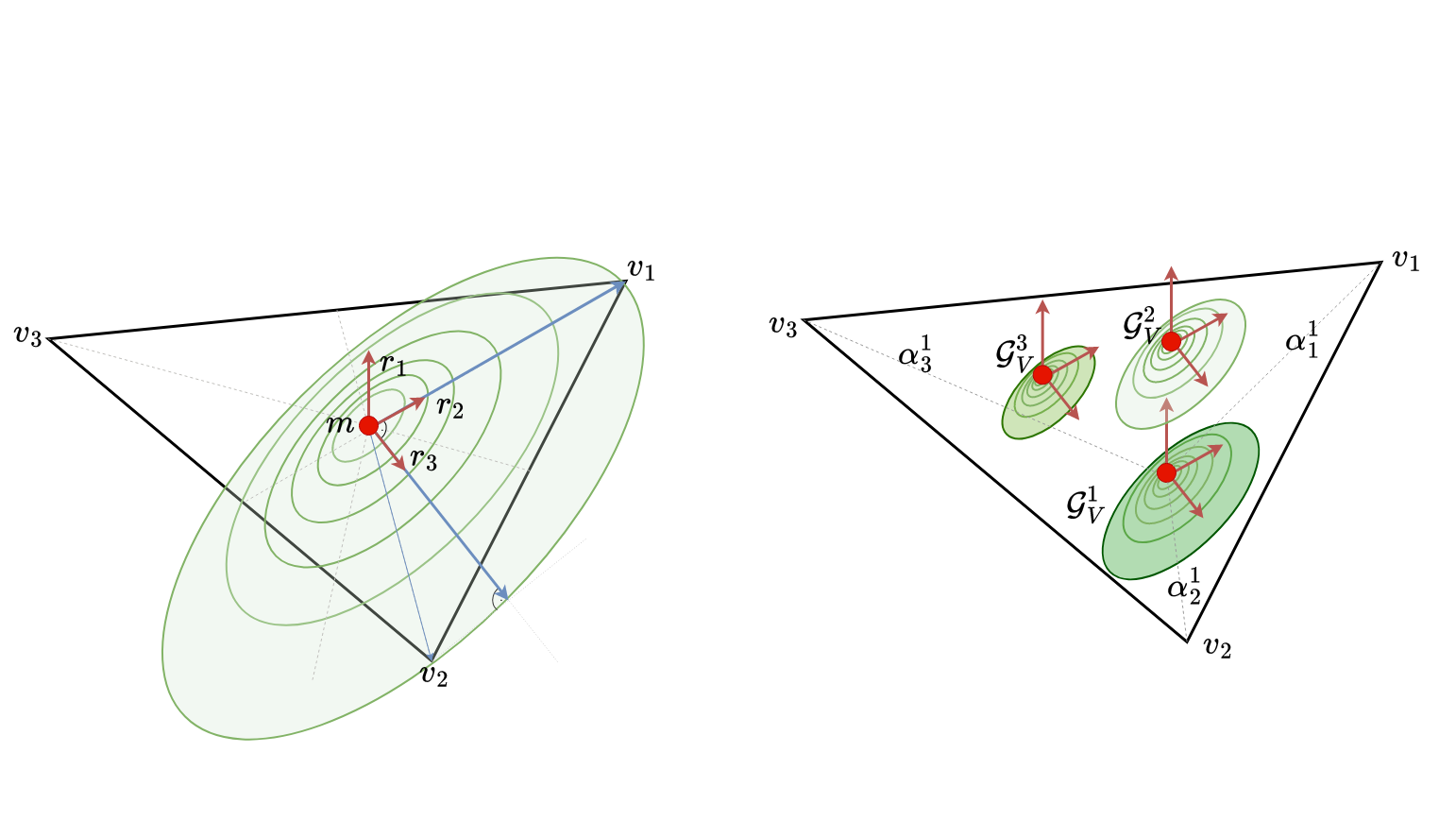}
 \caption{The left image presents the Gaussian component constructed on the face.  The right image presents how our model uses previously estimated Gaussian to construct $k$ (in Fig. $k=3$) components on the face.}
 \label{fig:splat_on_triangle}
\end{figure}

\paragraph{Gaussian Splatting (GS)}

The GS technique captures a 3D scene through an ensemble of 3D Gaussians, each defined by its position (mean), covariance matrix, opacity, and color. Additionally, spherical harmonics (SH) are employed to depict the color attributes. \cite{fridovich2022plenoxels, muller2022instant}.
The GS effectiveness is primarily attributed to the rendering process, which utilizes projections of Gaussian components. GS relies on a dense set of 3D Gaussian components with color and opacity: 
$$
\G = \{ (\N(\m_i,\Sigma_i), \sigma_i, c_i) \}_{i=1}^{n},
$$
where $\m_i$ denotes position, $\Sigma_i$ marks covariance, $\sigma_i$ stands for opacity, and $c$ is SH colors of $i$-th component.

In practice, GS uses factorization $\Sigma = RSS^TR^T$. We can train GS with a flat Gaussian distribution by forcing $s_1=0$, where $S=\mathrm{diag}(s_1,s_2,s_3)$.


The parameters of Gaussian distributions undergo direct training through gradient optimization. To enhance adaptability to complex scenes, GS employs additional training strategies. Notably, significant Gaussian components are subdivided. If the update parameters are substantial, Gaussian components are duplicated, which also can be removed due to their low transparency. 
In addition, the GS training procedure is implemented in the CUDA kernel, which supports fast training and real-time rendering.



\paragraph{Distribution on mesh faces}

In \our{} (baseline), we placed all Gaussian components on the mesh surface. Let us examine a single triangle face with vertices
$
V = \{ {\bf v}_1, {\bf v}_2, {\bf v}_3 \} \subset \R^3.
$
We aim to parameterize the Gaussian components using vertices from the face $V$. We express the mean vector as a convex combination of vertices $V$, thus determining the Gaussian splats positions:
\[
\m_{V}(\alpha_1,\alpha_2,\alpha_3) = \alpha_1  {\bf v}_1 + \alpha_2 {\bf v}_2 + \alpha_3 {\bf v}_3, 
\]
where $\alpha_1,\alpha_2,\alpha_3$ are trainable parameters such that \mbox{$\alpha_1 + \alpha_2 + \alpha_3 = 1$}. Through this parametrization, we consistently maintain the Gaussians positioned in the middle of face $V$.

Fig.~\ref{fig:affinie} shows examples of such transformation for rotation and scaling. Face $V$ at time zero corresponds to the triangle $V = \left\{{\bf v}^{V}_1, {\bf v}^{V}_2, {\bf v}^{V}_3\right\}$, then it is transformed into $V' = \left\{{\bf v}'^{V}_1, {\bf v}'^{V}_2, {\bf v}'^{V}_3\right\}$. Three Gaussians splats $\G_V^{1}$, $\G_V^{2}$, $\G_V^{3}$ depend only on the position of the vertices.

The covariance matrix can be defined as an empirical covariance calculated from three points. However, such a solution is not trivial when combined with the optimization proposed in GS. Instead, the covariance is parameterized by factorization:
$
\Sigma = RSS^TR^T,
$
where $R$ is the rotation matrix and $S$ the scaling parameters. Here, we define rotation and scaling matrices to stay in the original framework. 
Let start from orthonormal vectors: ${\bf r}_1,{\bf r}_2,{\bf r}_3 \in \R^3$, arranged into rotation matrix $R_V = [{\bf r}_1,{\bf r}_2,{\bf r}_3].$ Where the first vector is defined by the normal vector:
$$
\n = \frac{({\bf v}_2 - {\bf v}_1) \times ({\bf v}_3 - {\bf v}_1)}{\|({\bf v}_2 - {\bf v}_1) \times ({\bf v}_3 - {\bf v}_1) \|},
$$ 
where $\times$ is the cross product.
Given an explicitly defined mesh, we consistently possess knowledge of the vertex order for any given face. Hence,
to calculate ${\bf r}_2$,
we define the vector from the center to the vertex ${\bf v}_1$:
$$
{\bf r}_2 = \frac{ {\bf v}_1 - \m}{ \| {\bf v}_1 - \m \| },
$$
where 
$\m = \mathrm{mean}({\bf v}_1,{\bf v}_2,{\bf v}_3)$, which corresponds to the centroid of the triangle.

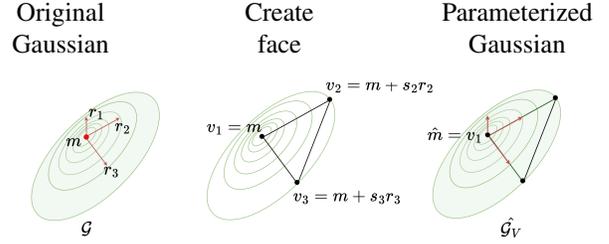
\begin{figure}[t]
\centering
  \begin{tikzpicture}
  \node [text width=1.5cm, align=center,] at (0, 0) {Original Gaussian };
\end{tikzpicture}
\hspace{1cm}
  \begin{tikzpicture}
  \node [text width=1.5cm, align=center,] at (1, 0) {Create face};
\end{tikzpicture} \hspace{1cm}
  \begin{tikzpicture}
  \node [text width=2cm, align=center,] at (1, 0) { Parameterized Gaussian}; 
\end{tikzpicture}\\
\vspace{0.25cm}
    \includegraphics[width=0.45\textwidth, trim=0 0 0 80, clip]{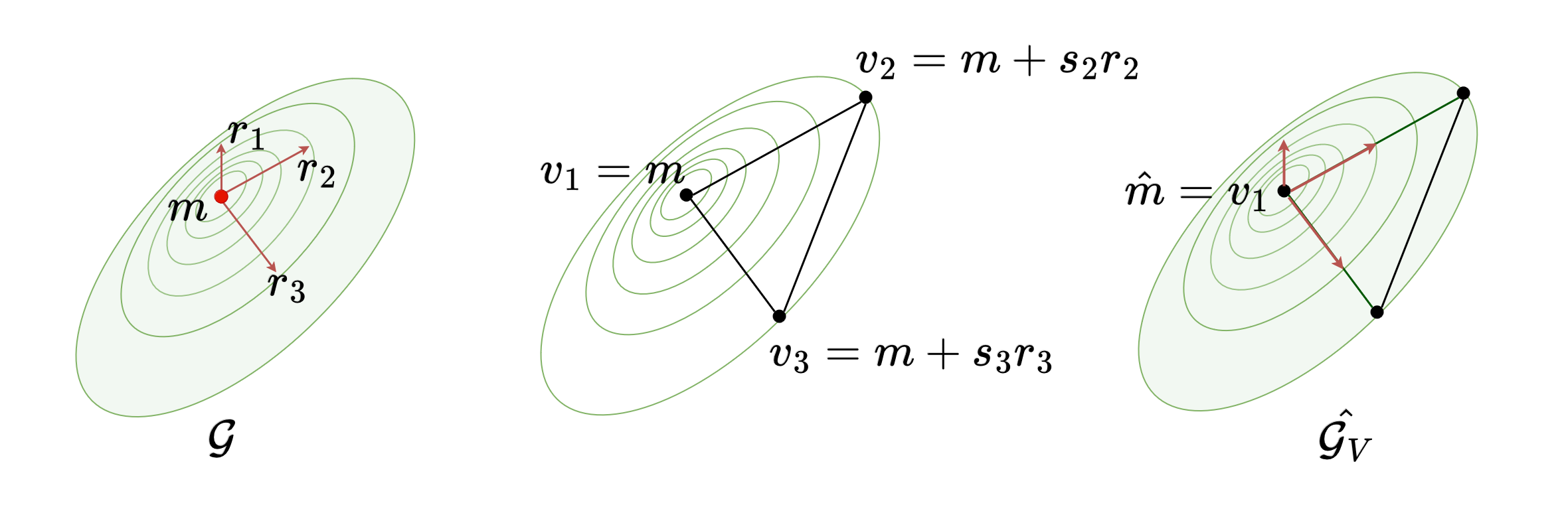}
\caption{Each Gaussian corresponds to one face from the \mesh{}. The model parametrizes the Gaussian depending on the vertices, allowing it to be edited.}
\label{fig:parametryzujG} 
\end{figure}

\begin{figure*}[t]
\qquad Only Gaussians means are on mesh \qquad \our{} \qquad \qquad \quad Only  Gaussians means are on mesh \qquad \qquad \our{} \\
    \includegraphics[width=0.49\textwidth]{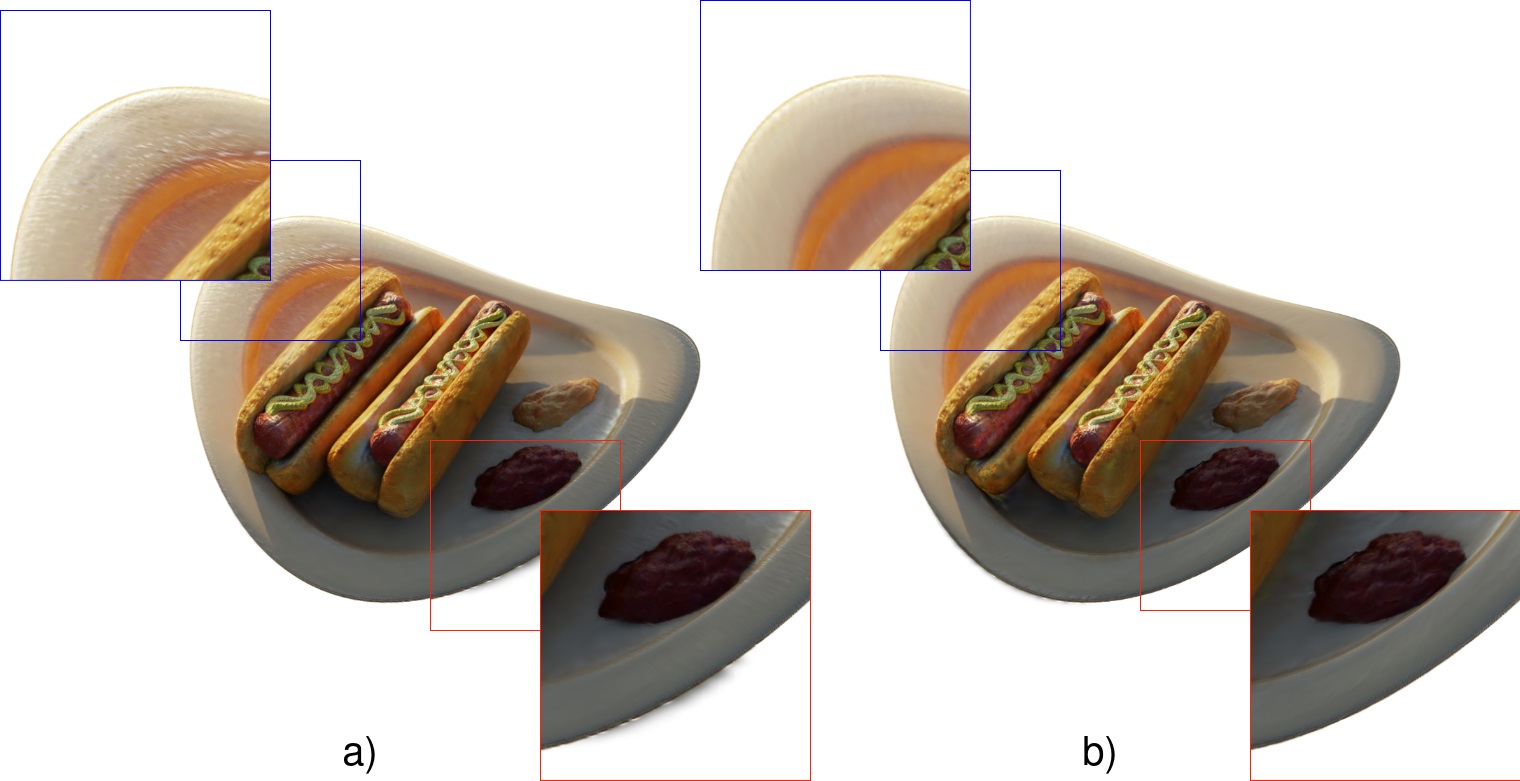}
    \includegraphics[width=0.49\textwidth]{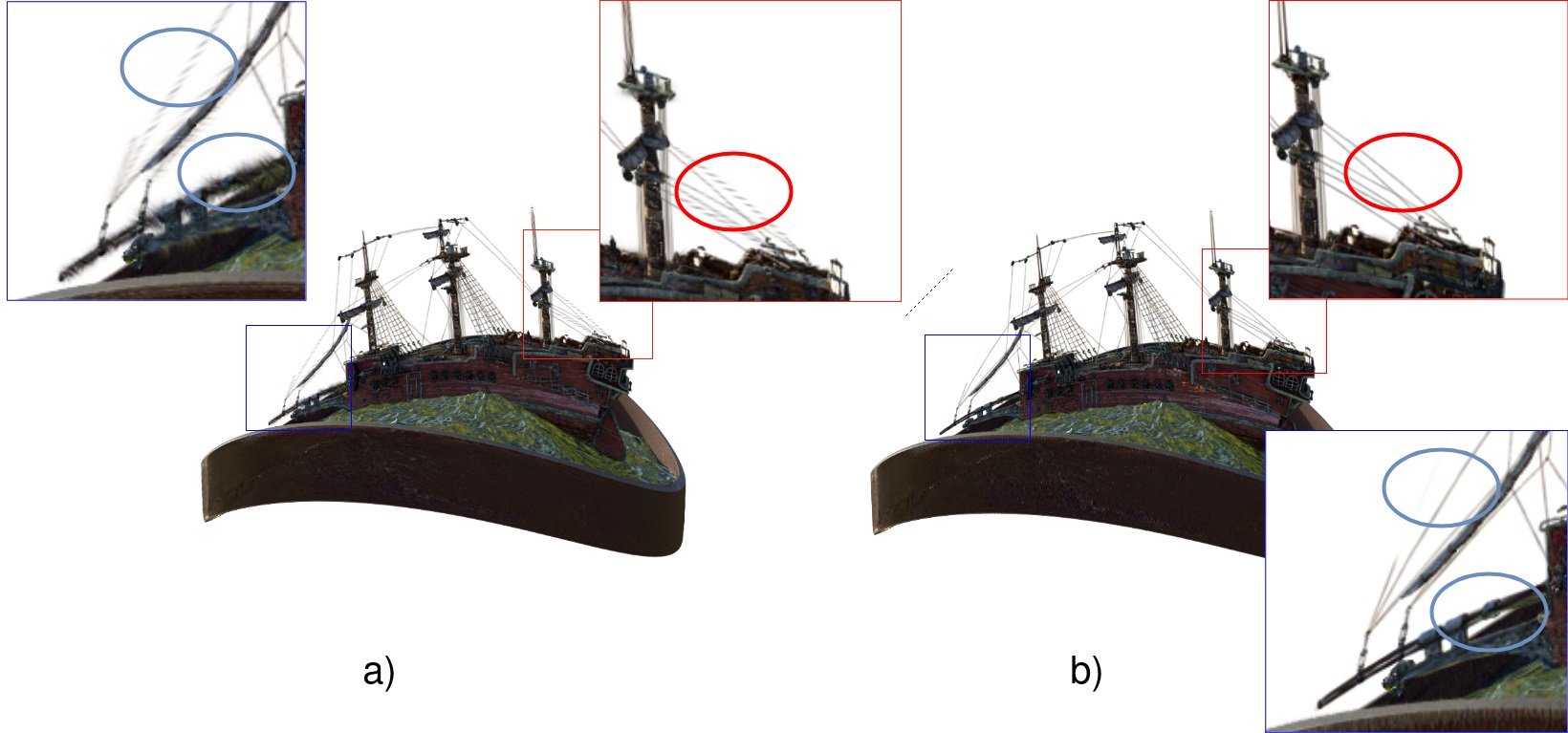}
\caption{
\our{} parametrizes all Gaussian components by mesh vertices to lie directly on the mesh. Therefore, when we modify the mesh positions, we automatically adapt the components' means and covariances. Consequently, we can see differences in the rendered image quality for the transforming when (a) only the Gaussian position is editable, and (b) when the Gaussian rotation and scaling are editable.}
\label{fig:ship_rotation} 
\vspace{-0.5cm}
\end{figure*}

The last vector is obtained through orthonormalizing the vector with respect to the existing two vectors (a single step in the Gram–Schmidt process~\cite{bjorck1994numerics}):
$$
\mathrm{orth}(\x;{\bf r}_1,{\bf r}_2) = x - \mathrm{proj}(\x,{\bf r}_1) - \mathrm{proj}(\x,{\bf r}_2), 
$$
where
$$
\mathrm{proj}({\bf v},{\bf u}) = \frac{ \langle {\bf v} , {\bf u} \rangle }{ \langle {\bf u},  {\bf u} \rangle } {\bf u}.
$$
To obtain ${\bf r}_3$, we use the vector from the center to the second vertex of the triangle:
$$
{\bf r}_3 = \frac{\mathrm{orth}({\bf v}_2 - \m ;{\bf r}_1,{\bf r}_2)}{ \| \mathrm{orth}( {\bf v}_2 - \m  ;{\bf r}_1,{\bf r}_2) \|}.
$$
In consequence, we obtain a rotation matrix 
$R_V = [{\bf r}_1,{\bf r}_2,{\bf r}_3]$, which aligns with the triangular face.
As scaling parameter $S$ we use
$
S_V = \mathrm{diag}(s_1,s_2,s_3),
$
where
$
s_1 = \varepsilon,
$
$
s_2 = \| \mathrm{mean}({\bf v}_1,{\bf v}_2,{\bf v}_3) - {\bf v}_1 \|,
$
and
$
s_3 = \langle {\bf v}_2 , {\bf r}_3 \rangle.
$
The first scaling parameters correspond with the normal vector. Since, in our case, the Gaussians are positioned on the faces, aligning with the surface~$s_1$ should be equal to zero. To avoid numerical problems, we fix this value as a small number (constant). 

Consequently, the covariance of Gaussian distribution positioned to face is given by:
$
\Sigma_{V} = R_VS_VS_V^TR_V^T,
$
and correspond with the shape of a triangle $V$. Fig. ~\ref{fig:splat_on_triangle} shows the process of determining Gaussians. Here the $\alpha_{j}^{i}$ refer to the $i$-th Gaussian $\G_V^{i}$ and the $j$-th vertex, respectively.

Within our model, we train a scale parameter, denoted as $\rho$, to dynamically adjust the Gaussian component size. For one face let us take $k \in \mathbb{N} $ number of Gaussians:
$$
\G_{V} = \{ \N(\m_V(\alpha^{i}_1,\alpha^{i}_2,\alpha^{i}_3),  \rho^{i} \Sigma_{V})  )  \}_{i=1}^{k},
$$
where
$\alpha^{i}_1 + \alpha^{i}_2 + \alpha^{i}_3 = 1$ and $\rho^{i} \in \R_{+}$. 

This distribution method enables us to render Gaussian splats solely dependent on the position of the mesh triangles. Consequently, we can readily perform geometric transformations when dealing with Gaussians within a single (triangle) face.
Moreover, triangle transformation will apply Gaussian transformation as well. 

The baseline demonstrates that objects can effectively be modeled using flat Gaussians and subsequently modified through a mesh-based representation.

A key limitation of our Baseline method is its dependence on an initial 3D mesh of the scene to parametrize flat Gaussians. To address this, we propose an approach that begins with a coarse or approximate mesh, such as the FLAME model, which we demonstrate in the experiments section. 

However, our model ultimately bypasses the need for any mesh by directly parameterizing Gaussians through a set of unconnected triangles—-a \mesh{}. This Mesh--Free approach allows for greater flexibility and reduces reliance on mesh accuracy, as each triangle independently defines the Gaussian parameters, achieving direct control over the representation.

\paragraph{Distribution using \mesh{}}
Without any pre-existing mesh, we can automatically generate \mesh{} structure, which can subsequently be used for object modification. We start by training GS with flat Gaussian splats, setting $s_1 = \epsilon$. Next, we estimate a triangle face for each Gaussian component and parameterize these components using the vertices. Notably, our \mesh{} is designed specifically to modify the GS model rather than to approximate object surfaces (see Fig.~\ref{fig:edit}). After re-parametrization, we obtain the final \our{}.


Let's assume Gaussian is defined by mean $\m$, rotation matrix $R=[{\bf r}_1,{\bf r}_2,{\bf r}_3]$ and scaling $S = \mathrm{diag}(\varepsilon,s_2,s_3).$
We define three vertex of a triangle:
$
V=[{\bf v}_1,{\bf v}_2,{\bf v}_3],
$
where ${\bf v}_1 = \m$, ${\bf v}_2 = \m+s_2  {\bf r}_2$, ${\bf v}_3 =\m+s_3{\bf r}_3$, see Fig. \ref{fig:parametryzujG}.

Now we freeze the vertices of the triangle $V=[{\bf v}_1,{\bf v}_2,{\bf v}_3]$ and re-parameterize the Gaussian component by defining $\hat \m$, $\hat R=[\hat {\bf r}_1,\hat {\bf r}_2,\hat {\bf r}_3]$ and
$\hat S = \mathrm{diag}(\hat s_1,\hat s_2,\hat s_3).$
First, we put \mbox{$\hat \m = {\bf v}_1$}. 
The first vertex of $\hat R$ is given by the normal vector:
$$
\hat {\bf r}_1 = \frac{({\bf v}_2 - {\bf v}_1) \times ({\bf v}_3 - {\bf v}_1)}{\|({\bf v}_2 - {\bf v}_1) \times ({\bf v}_3 - {\bf v}_1) \|},
$$
where $\times$ is the cross product.
The second one is defined by
$\hat {\bf r}_2 = \frac{({\bf v}_2-{\bf v}_1)}{\|({\bf v}_2-{\bf v}_1)\|}.$
The third one is obtained as a single step in the Gram–Schmidt process~\cite{bjorck1994numerics}: 
$$
\hat {\bf r}_3 = \mathrm{orth}({\bf v}_3-{\bf v}_1;{\bf r}_1,{\bf r}_2).
$$
Scaling parameters can also be easily calculated as:
$s_1= \varepsilon$, $\hat s_2 = \|{\bf v}_2-{\bf v}_1\|$ and  $\hat s_3 = \langle {\bf v}_3-{\bf v}_1, \hat {\bf r}_3 \rangle$. Consequently, the covariance of Gaussian distribution positioned to a triangle is given by
$
\hat \Sigma_{V} = \hat R_V \hat S_V \hat S_V^T \hat R_V^T,
$ 
and correspond with the shape of a triangle $V$. 
For one $V$, we define the corresponding Gaussian component:
$$
\hat \G_{V} =  \N(\hat \m_V, \hat \Sigma_{V})  ).  
$$

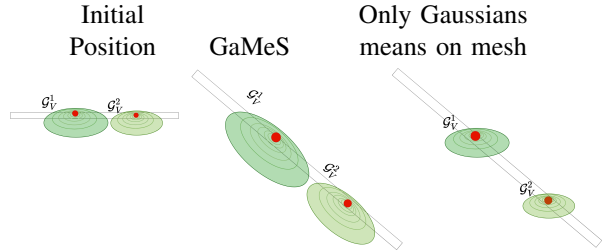
\begin{figure}[t]
 	\centering
  \begin{tikzpicture}
  \node [text width=1.5cm, align=center,] at (0, 0) {Initial Position};
\end{tikzpicture}
  \begin{tikzpicture}
  \node [text width=1.5cm, align=center,] at (4, 0) {\our{}};
\end{tikzpicture}
  \begin{tikzpicture}
  \node [text width=3cm, align=center,] at (2, 0) {Only Gaussians means on mesh };
\end{tikzpicture}\\
    \includegraphics[width=0.45\textwidth]{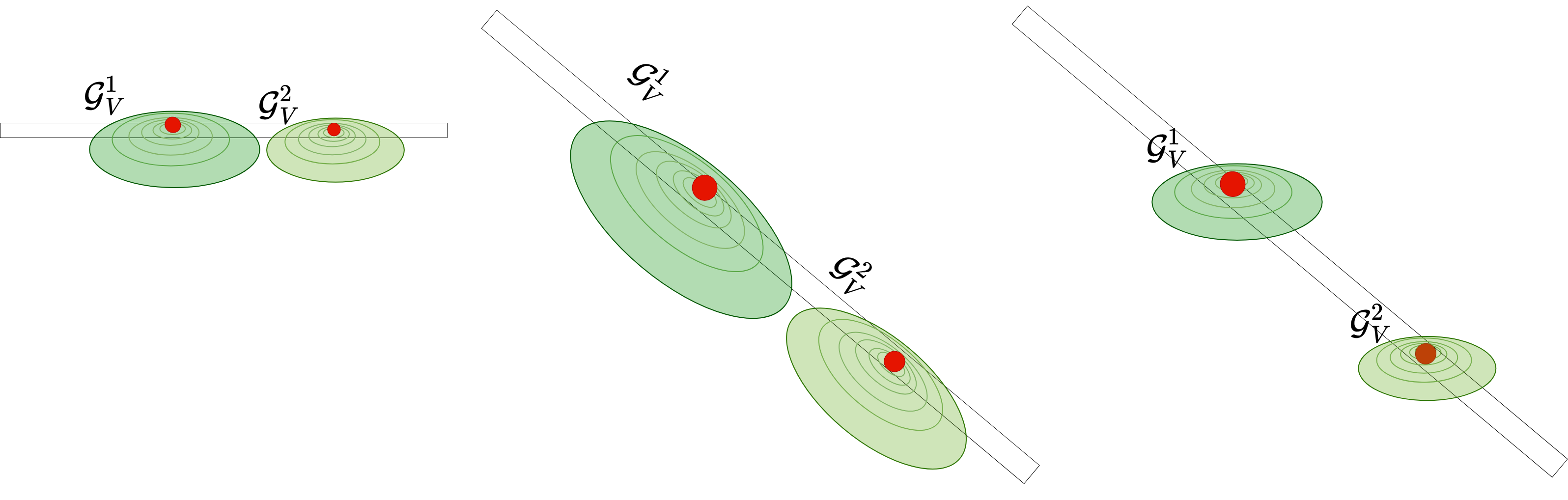}
\caption{
Compared to a method solely based on Gaussian averaging dependent on position, we observe that (following rotation and stretching the surface) Gaussians should dynamically adjust their scale and rotation to seamlessly conform to the change as in the case of \our{} model. }
    \vspace{-0.5cm}
\label{fig:rotation} 
\end{figure}

We obtain the Gaussian component parameterized by a triangle. 

\paragraph{\our{}: Gaussian Mesh Splatting}
The essential step in our model involves positioning Gaussians on the 3D mesh-based structure of the scene or object in question. Here, we have to consider two crucial experimental scenarios, namely, with (i.e., \textit{mesh-initialized input}) and without using mesh (i.e., \textit{mesh-free input}) during the training. 

First, we assume that a mesh is being provided and exclusively train the Gaussian components' centers, colors, and opacity. Here, numerous methods facilitate the generation of an initial mesh for objects, but our primary focus is not on this task. Instead, we emphasize addressing distinct issues that extend beyond the scope of initializing object meshes like EG3D~\cite{Chan2022}, NGLOD~\cite{takikawa2021nglod} or NeRFMeshing \cite{unknown}. Additionally, we also show the initialization of the mesh using \textit{Faces Learned with an Articulated Model and Expressions} (FLAME) \cite{FLAME:SiggraphAsia2017} can be used. 

Second, we consider the Mesh-Free Input scenario.
In the latter case, initialization of the mesh is required and achieved by utilizing our strategy with extracting \mesh{} from pre-trained GS. 

Let us consider a mesh-based structure denoted as: 
$
\mathcal{M} = (\mathcal{V},\mathcal{E}),
$ 
where  \mbox{$\mathcal{V} \subset \R^3$} represents vertices, and \mbox{$\mathcal{E} \subset \R^3$} signifies edges. The mesh's faces, denoted by: 
$$
\F_{\M} = \left\{ \left({\bf v}^{V_i}_1,{\bf v}^{V_i}_2,{\bf v}^{V_i}_3\right) \right\}_i^n
$$ 
are described by sets of vertices. As previously detailed, such mesh serves as a parametrization for shaping Gaussian components.


\begin{table}[h]
{\small 
\begin{center}
    \begin{tabular}{@{}l@{\;}l@{\;}l@{\;}l@{\;}l@{\;}l@{\;}l@{\;}l@{\;}l@{}}
         & \rot{ Chair } & \rot{Drums} & \rot{Lego} & \rot{Mic} & \rot{Materials} & \rot{Ship} & \rot{Hotdog} & \rot{Ficus} \\ 
 \hline

    \multicolumn{9}{c}{Static} \\ 
 \hline

NeRF & 33.00 & 25.01 & 32.54 & 32.91 & 29.62 & 28.65 & 36.18 & 30.13  \\ 
VolSDF & 30.57 & 20.43 & 29.46 & 30.53 & 29.13 & 25.51 & 35.11 & 22.91 \\ 
Ref-NeRF & 33.98 & 25.43 & 35.10 & 33.65 & 27.10 & 29.24 & 37.04 & 28.74\\ 
ENVIDR & 31.22 & 22.99 & 29.55 & 32.17 & 29.52 & 21.57 & 31.44 & 26.60 \\ 
  Plenoxels 
  & 33.98 & 25.35 & 34.10  & 33.26 & 29.14 &  29.62 & 36.43 & 31.83\\
GS & \bf 35.82 & \bf 26.17 & \bf 35.69 & \bf 35.34 & \bf 30.00 & \bf 30.87 & \bf 37.67 & \bf 34.83 \\ 
\hline
    \multicolumn{9}{c}{Editable} \\ 
 \hline
RIP-NeRF     & 34.84 & 24.89 & 33.41 & 34.19 &28.31 & 30.65 & 35.96 & 32.23\\
SuGaR & \bf  35.83 & \bf 26.15 & 35.78 & 35.36  & 30.00 & 30.80 & 37.72 & 34.87 \\
\our{} (our) &  35.38 & 26.03 & \bf 35.89 & \bf 37.16 & 29.62 & \bf 31.55 & 37.56 & 35.12\\
\our{} (our tr) & 34.74 & 25.94 & 35.88 & 36.58 & \bf 30.50 & 30.66 & \bf 37.84 & \bf 35.41 \\
  
\hline

    \end{tabular}
    \end{center}
    }
    \vspace{-0.3cm}
    \caption{Quantitative comparisons (PSNR) on a NeRF-Synthetic dataset showing that \our{} gives comparable results with other models. \our{} tr means \mesh{}.}
    \label{tab:nerf1}
\end{table} 

As the first scenario in our experiments, we choose the \mbox{number~$k \in \mathbb{N} $} of Gaussian components per mesh face and define how densely they should be placed on its surface. For mesh containing~\mbox{$n \in \mathbb{N} $} faces, the final number of Gaussians is fixed and equal to~\mbox{$k \cdot n$.} 

In the scenario with known mesh, for each Gaussian component, we define trainable parameters used to calculate mean and covariance $\{(\alpha^i_1, \alpha^i_2, \alpha^i_3, \rho^i)\}_{i=1}^{k \cdot n}$. Therefore, \our{} use a dense set $\G_{\M}$ of 3D Gaussians:
$$
\G_{\M} \!\! =\!\!\! \bigcup_{V_i \in \F_{\M}} \{ \G_{V_i}, \sigma_{i,j}, c_{i,j}  \}_{j=1}^{k},
$$
where $\sigma_{i,j}$ is opacity, and $c_{i,j}$ is SH colors of $j$-th Gaussian, and $i$-th face. 

When we do not have mesh, we train GS, and for each Gaussian, we estimate a triangle. Then we obtain set $\G_{\M}$ of 3D parameterized Gaussians $\hat \G_V$ on \mesh{} $\hat{\M}$:
$$
\G_{\M} \!\! =\!\!\! \bigcup_{V_i \in \F_{\hat{\M}}} \{ \hat \G_{V_i}, \sigma_{i}, c_{i}  \},
$$
where  $\sigma_{i}$ is opacity, and $c_{i}$ is SH colors Gaussian corresponded to $i$-th face. 

During training, we can modify mesh vertices and Gaussian parameters. We can also use external tools for mesh fitting~\cite{takikawa2021neural} and train only Gaussian components. In the experimental section, we show both of these scenarios. 


\paragraph{Mesh modification}

\our{} parameterize all Gaussian components by vertices of mesh in such a way that they lie on meshes (or define \mesh{}). Therefore, when we modify mesh/\mesh{}, we automatically adapt the means and covariances of relevant Gaussians. Such an approach allows us to work with transformation. To better visualize such properties, we compare our model with a solution where the Gaussian components' centers are placed on mesh faces but do not parameterize the covariance matrix. In Fig.~\ref{fig:ship_rotation}, we show that \our{} Gaussian component adapts automatically to mesh modification.

In particular, when an object is bent, models that would only consider the position (means) of Gaussians depending on the mesh would fail to adapt properly. We can observe this effect in Fig. \ref{fig:ship_rotation} with the ship's rope rendering. In contrast, \our{} provides a perfect fit with the modification, and the operation scheme is shown in Fig. \ref{fig:rotation}.

\section{Experiments}

    


\begin{figure}[h]
     	\centering
    \includegraphics[width=0.45\textwidth, trim=0 0 0 0, clip]{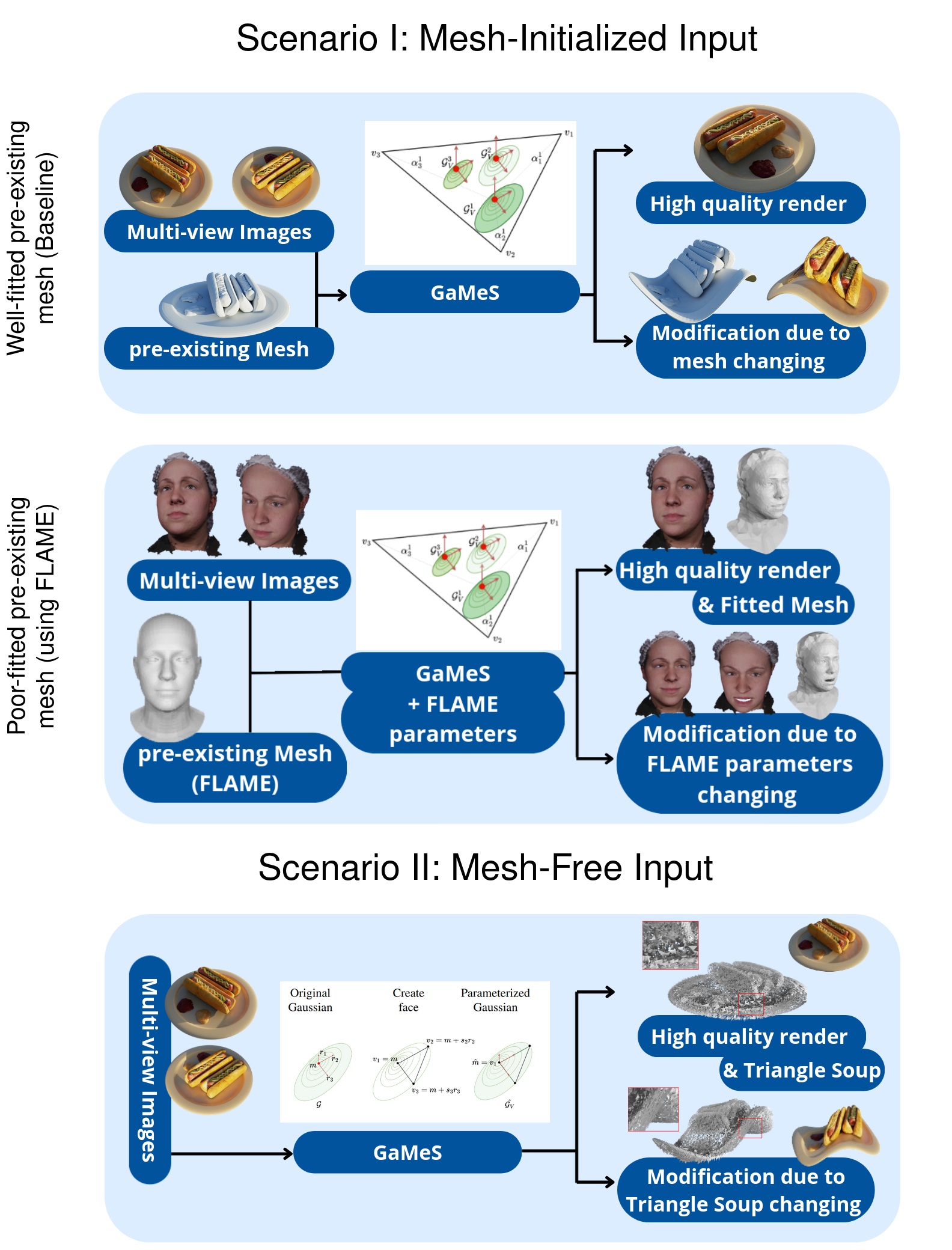}
\caption{A diagram illustrating the model, covering two scenarios: (1) with a pre-existing mesh, which may either be well-fitted and left unmodified during training (baseline) or serve as a rough initialization like FLAME model that is refined throughout training, and (2) without an initial mesh or mesh estimation provided as input.}
\label{fig:scenariusz} 
\end{figure}

Here, we delineate implementation details and describe the used datasets and the reasons for their selection. In our experiments, we consider two distinct scenarios, i.e., \textit{mesh-initialized input}, and \textit{mesh-free input} (see Fig. \ref{fig:scenariusz}).

\paragraph{Implementation Details and Data Description}
The time taken by the optimization process for our model varied from a few up to no more than sixty minutes, as it depends on factors like scene complexity, dataset characteristics, the number of Gaussians per face, and vertex fine-tuning decisions. Nonetheless, all experiments were conducted within a reasonable time frame, underscoring the efficiency of our approach across diverse scenarios. The supplementary material provides all the relevant details, and the source code is available in GitHub\footnote{\url{https://github.com/waczjoan/gaussian-mesh-splatting}}. We illustrated the fundamental gains of \our{} model through experiments with three distinct datasets. These datasets are as follows.

\begin{table}[h]
{\small 
\begin{center}
    \begin{tabular}{@{}l@{\;}c@{\;}c@{\;}c@{\;}c@{\;}c@{\;}c@{\;}c@{\;}c@{\;}c@{}}
    &  \multicolumn{5}{c}{Outdoor scenes} & \multicolumn{4}{c}{Indoor scenes} \\
    & \rot{bicycle} & \rot{flowers} & \rot{garden} & \rot{stump} & \rot{treehill} & \rot{room} & \rot{counter} & \rot{kitchen} & \rot{bonsai} \\ \hline

    \multicolumn{10}{c}{ Static } \\ \hline
     
    INGP   & 22.17 & 20.65 & 25.07 & 23.47 & 22.37 & 29.69 & 26.69 & 29.48 & 30.69 \\
    M-NeRF & 24.37 & \bf 21.73 & 26.98 & 26.40 & \bf 22.87 & \bf 31.63 & \bf 29.55 & \bf 32.23 & \bf 33.46 \\
    GS 7K & 23.60 & 20.52 & 26.25 & 25.71 & 22.09 & 28.14 & 26.71 & 28.55 & 28.85\\
    GS 30K  & \bf 25.25 & 21.52 & \bf 27.41 & \bf 26.55 & 22.49 & 30.63 & 28.70 & 30.32 & 31.98 \\
    \hline
        \multicolumn{10}{c}{ Editable } \\ \hline
    SuGaR  & 23.14 & - & 25.36 & 24.70 & - & 30.03 & 27.62 & 29.56 & 30.51 \\
    \our{} & \bf 24.99 & \bf 21.27 & \bf 27.22 & \bf 26.54 & \bf 22.39 & \bf 31.52 & \bf 28.92 & \bf 31.12 & \bf 32.09
    \\ \hline
    \end{tabular}
    \end{center}
    }
    \vspace{-0.3cm}
    \caption{The quantitative comparisons of reconstruction capability (PSNR) on Mip-NeRF360 dataset. R-SuGaR-15K uses 1M vertices. \our{} uses \mesh{}.   }
    \vspace{-0.3cm}    
    \label{tab:mipnerf360psnr}
\end{table}



{\em Synthetic Dataset}: 
A NeRF-provided eight geometrically complex objects with realistic non-Lambertian materials \cite{mildenhall2020nerf}. In addition to containing images of the objects, the dataset also incorporates corresponding meshes. We leveraged this dataset for our initial experiments to unequivocally showcase the capabilities of our model and underscore its flexibility in mesh editing.

{\em Mip-NeRF360 Dataset}:
A dataset comprising five outdoor and four indoor scenes, each featuring intricate central objects or areas against detailed backgrounds \cite{barron2022mipnerf360}. 
We used this dataset to demonstrate the adaptability of our model in handling dense scenes, employing a non-conventional strategy during the initial training stage in mesh preparation.

{\em Human Faces Dataset}:
A modified subset of the Stirling/ESRC 3D Face Database\footnote{\url{https://pics.stir.ac.uk/ESRC/index.htm}}, that includes images of six people, generated using \textit{Blender} software from various perspectives, consistently excluding the backs of heads. Only one expression is assigned for each face. Notably, the dataset lacks corresponding mesh representations for the depicted person. We used this dataset to compare \our{} with the NeRFlame model \cite{zajac2023nerflame}, mainly within the human-mesh fitting task.

\paragraph{Scenario I: mesh-initialized input}

\begin{figure}[h]
 	\centering
    \begin{tabular}{cc}
Original mesh  &  Subdivide larger faces \\
    \includegraphics[width=0.22\textwidth, trim=0 50 0 100, clip]{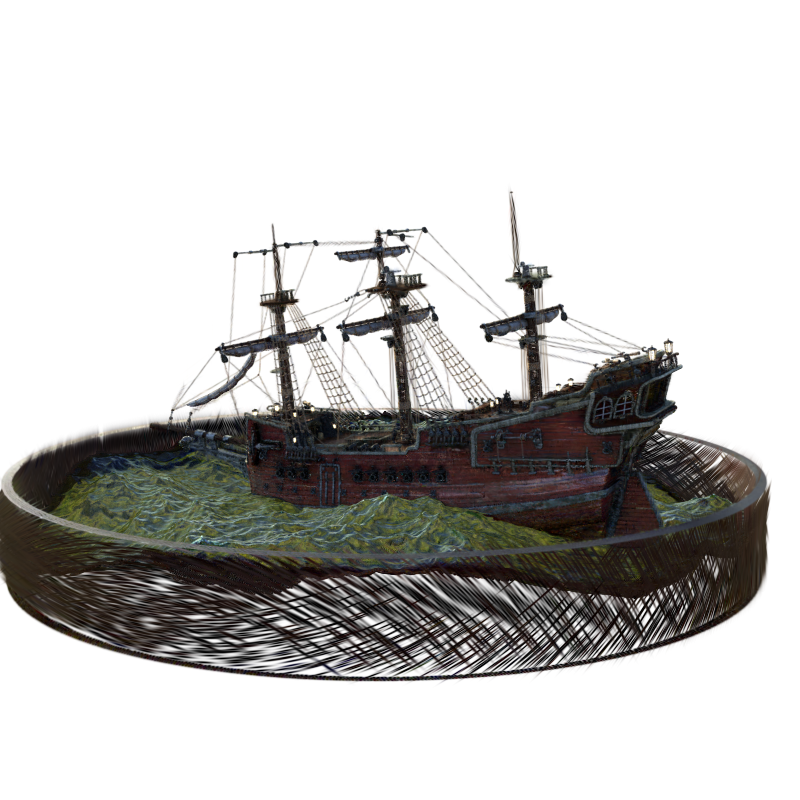} &
    \includegraphics[width=0.22\textwidth, trim=0 50 0 100, clip]{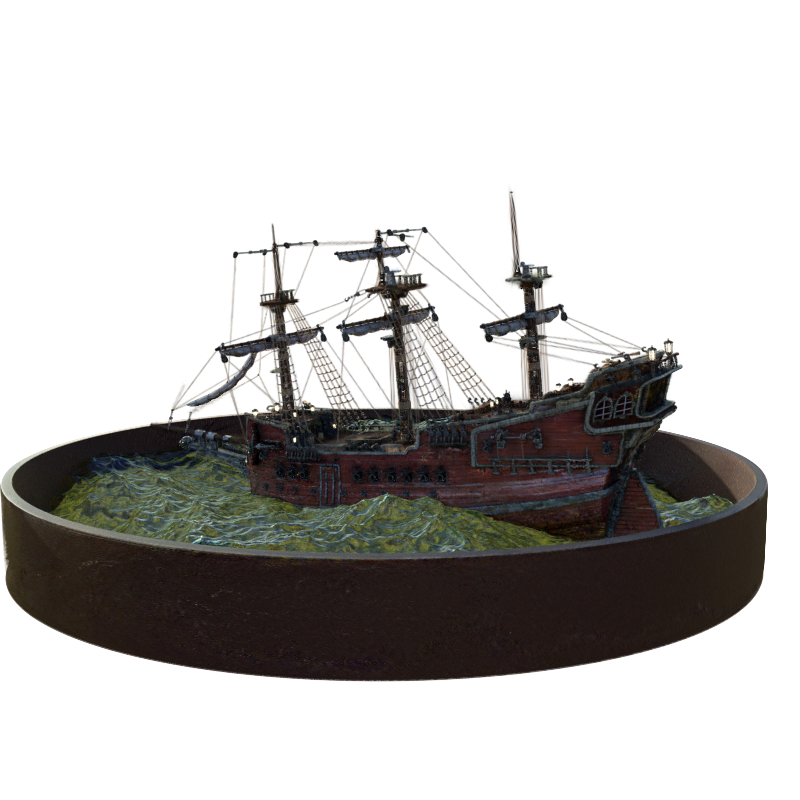}
	\end{tabular}
\caption{\our{} (baseline) uses a fixed number of faces. In the case of a mesh containing faces of different sizes, finding such parameters causes problems. 
In such cases, we propose to divide large faces into smaller parts. The left side includes only the original mesh, while on the right, sizable mesh faces of the brown pool surrounding have been subdivided into smaller ones.}
\label{fig:comparisonsubdivide} 
\end{figure}

In this scenario, we utilize the provided mesh, incorporating Gaussians by strategically placing them onto its surface. We also allowed for moving vertices. In implementation, we use a fixed number of Gaussian components per face. In the case of models with various face sizes, finding such parameters is not trivial. To solve such a problem, we propose to divide large faces into smaller parts, as shown in Fig. \ref{fig:comparisonsubdivide}.

{\em Well-fitted pre-existing mesh (baseline):} \label{sec:nerf}
Our initial experiments used the Synthetic Dataset \cite{mildenhall2020nerf}, incorporating shared meshes. Table \ref{tab:nerf1} presents PSNR metrics demonstrating the competitiveness of \our{} approach with existing methods. We provide additional numerical comparisons in the supplementary material. Notably, a significant contribution lies in the ease with which renders can be modified manually or in an automated process. Fig.~\ref{fig:nerfsyntheticslego} shows examples of reconstruction and simple object animations. 

\begin{table}[H]
{\small 
\begin{center}
    \begin{tabular}{lllllll}
     Person &\hspace*{-0.2cm} 1&\hspace*{-0.2cm} 2&  \hspace*{-0.2cm} 3&\hspace*{-0.2cm} 4&\hspace*{-0.2cm}5&\hspace*{-0.2cm} 6 
 \\ \hline

\multicolumn{7}{c}{Static}
 \\ \hline
 
\arrayrulecolor{lightgray}\hline \arrayrulecolor{black}
 NeRF  & 33.37 & 33.39 &33.08 &31.96 & 33.15 &32.42  \\
 GS &\bf 50.16& \bf 49.04& \bf 49.81& \bf 49.97&\bf 49.77&\bf 49.17\\\hline

\multicolumn{7}{c}{Editable}
 \\ \hline
  NeRFlame  & 27.89 & \bf 29.79  & \bf 29.70 & 25.78  & \bf 32.59 & 29.18   \\
\our{}&  \bf 32.73 & 29.56 & 29.42 & \bf 29.95 & 32.27  & \bf 31.50 \\\hline
    \end{tabular}
    \end{center}
    }
    \caption{ Comparison of PSNR, SSIM, and LPIPS matrices between static models: NeRF, GS and referring to them: NeRFlame and \our{}. We used 100 Gaussians per face.}
    \label{tab:facescomparisionnerflame}
\end{table} 

\begin{figure}
\includegraphics[width=0.48\textwidth]{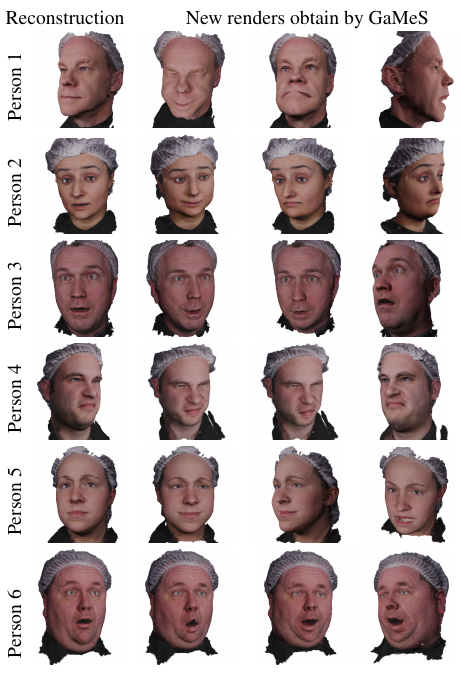}

\caption{Reconstruction of 3D peoples and generation of new expressions using the \our{} model.}

\label{fig:reconst} 
\end{figure}

{\em Poor-fitted pre-existing mesh: \our{} with FLAME:}
In this experiment, we show the possibility of fitting mesh using a pre-existing mesh, which was not acquired directly from our data in contrast to previous sub-scenarios. Here, we relied on the Faces Dataset as it provides the required experimental data. To initialize the meshes, we used the FLAME \cite{FLAME:SiggraphAsia2017} framework, generating fully controllable meshes of human faces. Consequently, we could highlight the critical advantage of \our{} model, i.e., the ability to modify objects. We used the official implementation of the FLAME, with the number of parameters suggested in RingNet \cite{RingNet}
In addition, we implemented an extra detail learning parameter to better represent hair or T-shirts.



Initially, a human mesh is generated using the FLAME model. By utilizing the FLAME parameters, we can modify various aspects, especially facial expressions, like smiling, sadness, eye closure, mouth opening, and flirty or disgusted expressions. The possibilities of reconstruction and modifications for all six faces are depicted in Fig. \ref{fig:reconst}. All expressions were chosen randomly as our main goal was to show \our{} model's ability to render objects as well as its modifications/animations.

Table~\ref{tab:facescomparisionnerflame} shows the results of using PSNR metrics on six faces. We made the comparisons between the two baseline NeRF models, GS, and the models created from them, i.e., the corresponding NeRFlame and \our{} with animation capabilities.

\paragraph{Scenario II: mesh-free input}
When we do not have mesh, we train GS with flat Gaussians. Then, we construct the \mesh{}. In Table \ref{tab:nerf1}, we show a comparison with other models on the Synthetic Dataset. Our approach is also capable of modeling unbounded scenes. In order to achieve that, we first create an initial \mesh{} and then adjust it during the training. Results in Table~\ref{tab:mipnerf360psnr} demonstrate that such modeling is feasible and reasonably comparable to other existing models. Particularly, the results show parity with the GS-30K and R-SuGaR-15K models. 

In this work, we demonstrate the editability of \mesh{}. Fig. \ref{fig:nerfsyntheticslego} in the first and second rows show that edits to the ficus's or excavator's \mesh{} can yield highly realistic modifications, whereas the third row illustrates hotdog's edits that are unrealistic. In Fig. \ref{fig:edit}, we explore three distinct approaches to editing a Ficus model. The first method involves direct modifications to the \mesh{}, while the second and third methods employ deformations via an estimated mesh, as proposed in \cite{waczynska2024dmisoeditingdynamic3d}. Meshes were scaled and structured using the AlphaShape algorithm~\cite{1056714}, facilitating a comparison between realism and edit complexity. We observe that increased accuracy in mesh estimation enables more intricate edits, further expanding the range of achievable modifications.

\section{Conclusion}

By integrating these two concepts, we enable a structured approach to scene modeling that connects seamlessly to a pre-existing mesh. Our results demonstrate that the \mesh{} is well-suited for replacing conventional meshes in animation contexts, allowing a flexible parametrization of a \mesh{} based on the estimated mesh. Leveraging the capabilities of the \our{} model, we can parametrically control flat Gaussians, facilitating intuitive and precise real-time object editing.


\textbf{Limitations} \our{} allows real-time modifications, but artifacts appear in case of significant changes in the case of meshes with large faces. In practice, large faces should be divided into smaller ones. How to change Gaussian components in \our{} when mesh faces are split is not apparent. 

\textbf{Social Impact} 
The model has the potential to significantly enhance object modification in games and virtual reality, offering exciting applications in simulations, particularly those involving physics-based interactions.





\section{Appendix}

The appendix to the paper contains section explaining benefits of \mesh{} as well as experiments related to NeRF Synthetic, Faces dataset and a continuation of the numerical results expounded in the main document

\subsection{Advantages of \our{} over vanilla Gaussian Splatting}
\begin{figure}[h!]
    \centering
    \begin{subfigure}{.24\textwidth}
      \centering
      \includegraphics[width=.9\textwidth]{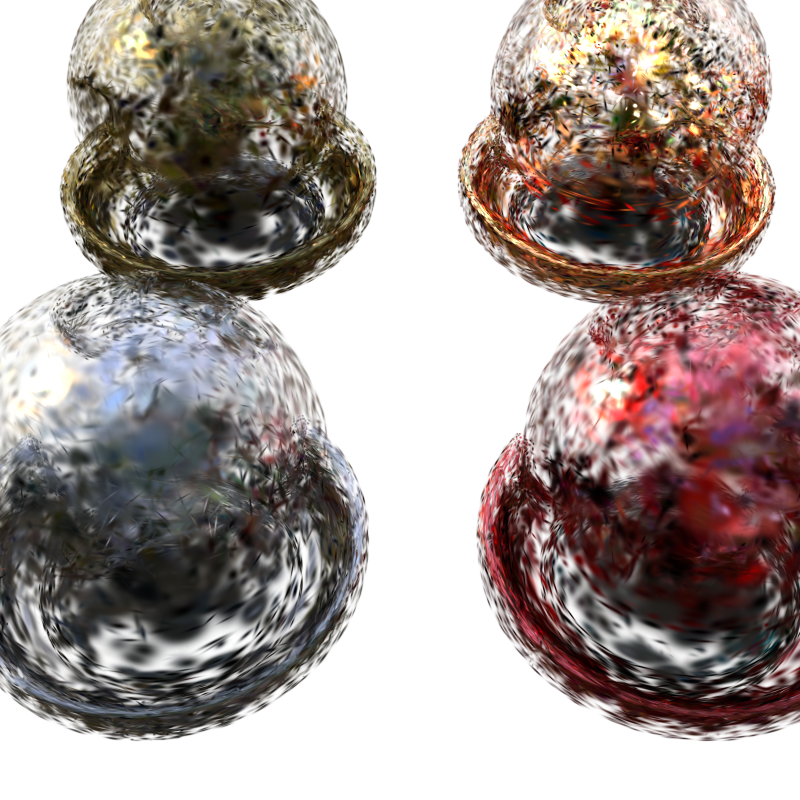}
    \end{subfigure}%
    \begin{subfigure}{.24\textwidth}
      \centering
      \includegraphics[width=.9\textwidth]{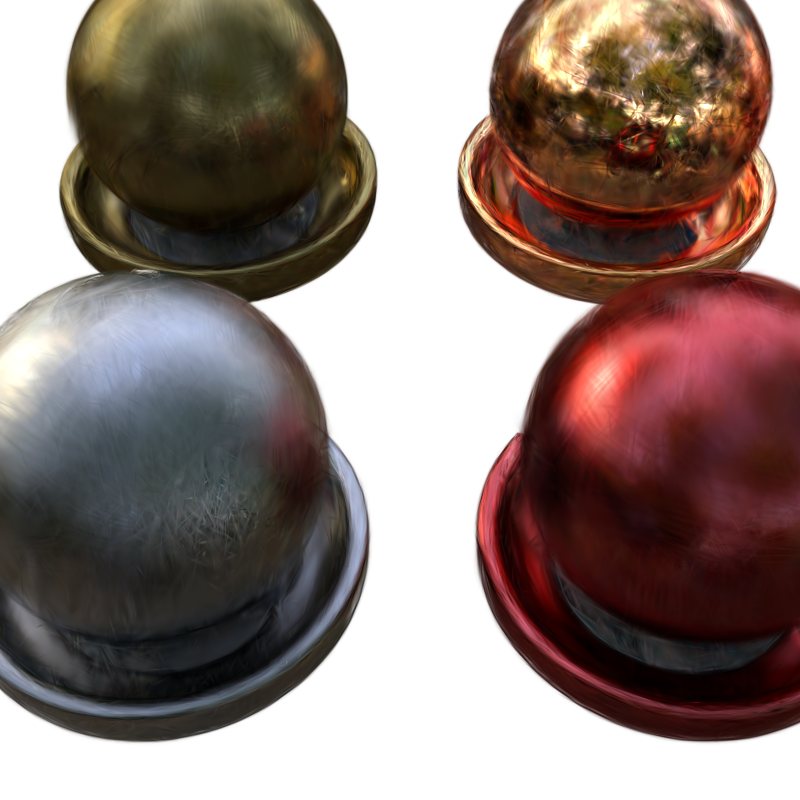}
    \end{subfigure}
    \\
    \begin{subfigure}{.24\textwidth}
      \centering
      \includegraphics[width=.9\textwidth]{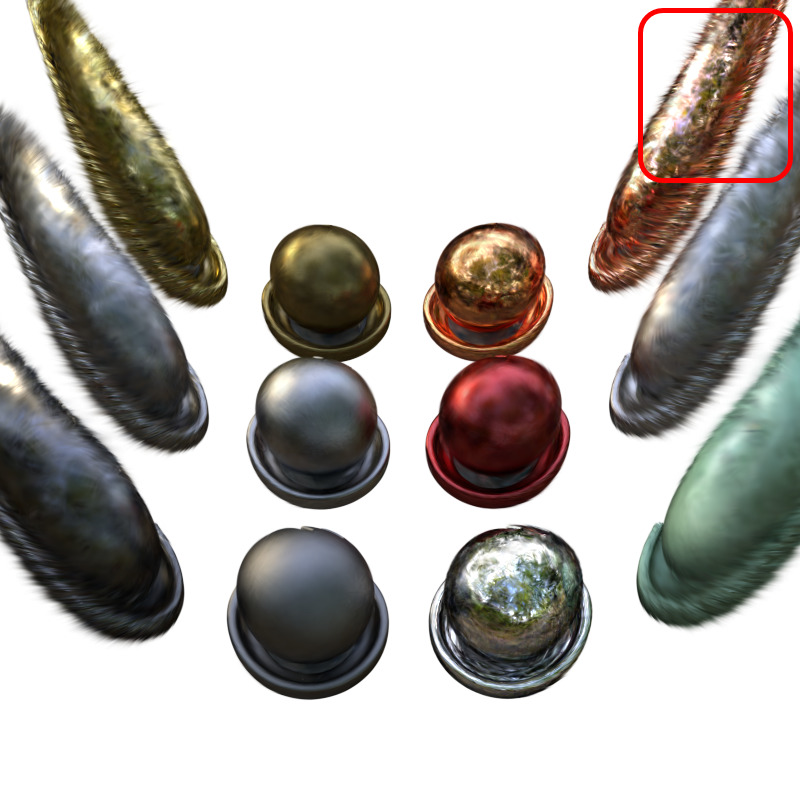}
      \caption{Gaussian splatting}
    \end{subfigure}%
    \begin{subfigure}{.24\textwidth}
      \centering
      \includegraphics[width=.9\textwidth]{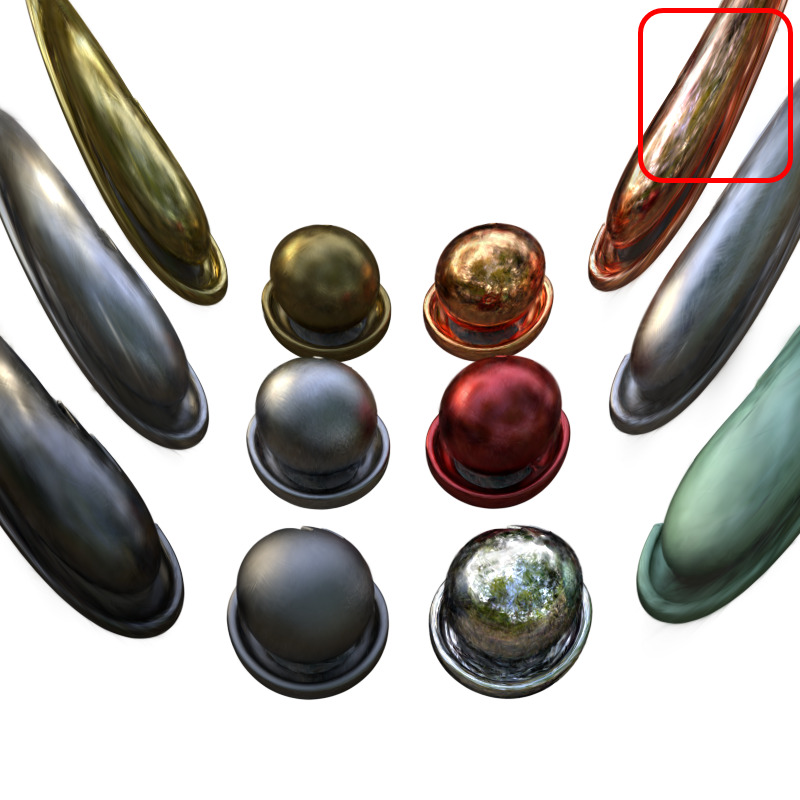}
      \caption{\our{}(our)}
    \end{subfigure}%
    \caption{Examples of different modifications, where updating covariance matrix associated with guassians is necessary. In the case of Gaussian splatting one can see that holes and some blurriness might appear. On the other hand, our method automatically recalculates the parameters of gaussian splats when the vertices are moved. The first row shows version of the original image with rescaled positions of means and the second image shows translation and rotations on the sides of an image.}
    \label{fig:rescale_comp}
\end{figure}

When modifying the object represented by Gaussian splats, each Gaussian covariance matrix must be updated to reflect the changes made to the object correctly. An example of a modification without changing the covariance matrix can be seen in Fig. \ref{fig:rescale_comp}. Such changes can be related to rescaling the covariance or applying an additional rotation. We argue that our approach allows for an easier updating of the Gaussians.

Although it is possible to rescale each Gaussian splat by updating the matrix $S$ in the decomposition of a covariance matrix given by $\Sigma = RSS^TR^T$, this can be done by calculation $S'=AS$, where $A\in\R^{3\times3}$ is a diagonal matrix with scaling factors on the diagonal. The new covariance matrix can be calculated as $\Sigma' = RS'S'^TR^T = RASS^TA^TR^T$. However, the exact values of the scaling factors are not always apparent for the intended modification. For example, this can be the case when the scale is different for each Gaussian. On the other hand, our approach allows for the movement of the triangle vertices directly as a point in $\R^3$ using some graphical interface (e.g., blender). The covariance matrix is automatically recalculated for the updated Gaussians, see Fig. \ref{fig:rescale_comp}.

A similar argument can be made for the rotation of a Gaussian splat. The new rotation can be calculated by substituting $R'=BR$, where $B\in\text{SO}(3)$. The covariance matrix can be calculated as $\Sigma' = R'SS^TR'^T = BRSS^TR^TB^T$. In our approach, this can be done by moving the points in $\R^3$, and the new rotation will be calculated automatically, see Fig. \ref{fig:rescale_comp}.

\section{Implementation details}

In the main paper, we evaluated the models and presented the results on a black background.

\begin{figure}
 	\centering
	 \includegraphics[width=0.5\textwidth, clip]{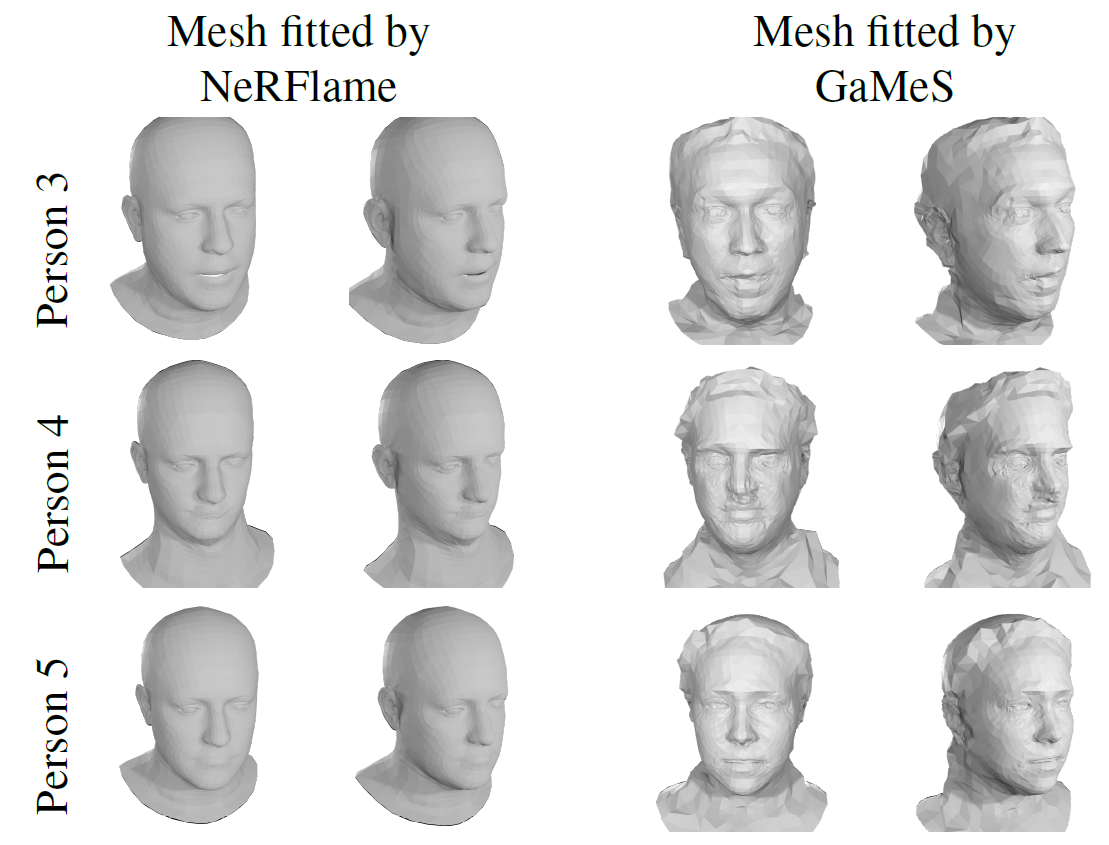}
\caption{During the training of \our, we simultaneously model FLAME mesh. Here, we present the meshes fitted by \our{} compared to the fitted mesh by NeRFlame model.}
\label{fig:mesh} 
\vspace{-0.5cm}
\end{figure}

\subsection{Limitations and future improvement}

\begin{table*}[t]
{\small 
\begin{center}
    \begin{tabular}{l|cccccccc}
    \multicolumn{9}{c}{PSNR $\uparrow$} \\    
Number of  & Chair & Drums & Lego & Mic & Materials & Ship & Hotdog & Ficus \\
Gaussians &  &  &  &  &  &  &  &  \\
\hline
1  & 29.32 & 25.18 & 32.74 & 32.48 & 25.37 & 27.45 & 32.10 & 32.05 \\
5 & 30.70 & 25.35 & 33.21& 32.49 & 25.29 & 29.29 & 33.40 &32.29\\
10 &31.11 & 25.36 & OOM & 32.29 & 25.25 & 29.56&  33.65& 32.35 \\
\mesh{} & \bf 35.34 & \bf 25.93 & \bf 35.42 & \bf 35.34 & \bf 30.01 & \bf 29.58 & \bf 37.65 & \bf 34.81 
\\ 
    \multicolumn{9}{c}{SSIM $\uparrow$} \\
& Chair & Drums & Lego & Mic & Materials & Ship & Hotdog & Ficus \\ \hline
1  & 0.934 & 0.940 & 0.972 & 0.984 & 0.916 & 0.879 & 0.959 & 0.978\\
5 & 0.950 & 0.942 & 0.975 & 0.984 & 0.914 & 0.894 & 0.968 & 0.978\\
10 & 0.951 & 0.942 & OOM & 0.983 & 0.913 & \bf 0.896 & 0.970 & 0.979\\
\mesh{} & \bf 0.986 & \bf 0.952 & \bf 0.982 & \bf 0.991 & \bf 0.958 & 0.886 & \bf 0.985 & \bf 0.987 \\
    \multicolumn{9}{c}{LPIPS $\downarrow$ }\\ 
& Chair & Drums & Lego & Mic & Materials & Ship & Hotdog & Ficus \\ \hline
1  & 0.066 & 0.052 & 0.027 & 0.012 & 0.060 & 0.133 & 0.062 & 0.018\\
5 & 0.050 & 0.049 & 0.022 & 0.012 & 0.061 & 0.107 & 0.042 & 0.018\\
10& 0.044 & 0.049 & OOM & 0.012 & 0.063 & \bf 0.102 & 0.038 & 0.019 \\
\mesh{} & \bf 0.012 & \bf 0.039 & \bf 0.015 & \bf 0.006 & \bf 0.033 & 0.105 & \bf 0.019 & \bf 0.012  
    \end{tabular}
    \end{center}
    }
    \caption{The quantitative comparisons (PSNR / SSIM / LPIPS) on NeRF-Synthetic dataset with original mesh not moving during training meshes or \mesh{}, with white background. OOM - CUDA out of memory. In this experiment, we used Tesla V100-SXM2-32GB GPU.}
    \label{tab:ablationsyntetic}
\end{table*}

As we refrain from employing a pruning algorithm or any other mechanism to regulate the number of Gaussians in our experiments aimed at achieving optimal results within a given mesh (
Well-fitted pre-existing mesh (baseline) in the main paper), we opted to utilize meshes with a higher facet count than those present in Nerf Synthetic. To achieve this, we leveraged Blender's Subdivide option, which involves splitting selected edges and faces, introducing new vertices, and appropriately subdividing the implicated faces, leading to markedly improved results. While this case did not involve numerous large faces, we recognize the potential challenges it may pose. Noticeable gaps are evident in this area. Adapting the Gaussian count based on face size is undoubtedly an area for further development.

In contrast, for the ablation study, we adhered to the use of the originally prepared meshes to facilitate reproducible. However, we emphasize the importance of a good choice of face surface, since all Gaussians align perfectly with surface. 

A prospective avenue for advancing the project involves the implementation of an automated approach for selecting the maximum and minimum face mesh areas. Such a solution would pave the way for a more versatile and generic mode. 

\begin{figure}[h]
 	\centering
    \begin{tabular}{ccc}
    \multicolumn{2}{c}{Initial Mesh} & Subdivided Mesh\\

$k = 1$ &  $k = 10$ & $k = 5$ \\
PSNR = 29.32 & PSNR =  31.11 & PSNR =  33.01 \\


\includegraphics[width=0.12\textwidth, trim=150 0 150 120, clip]{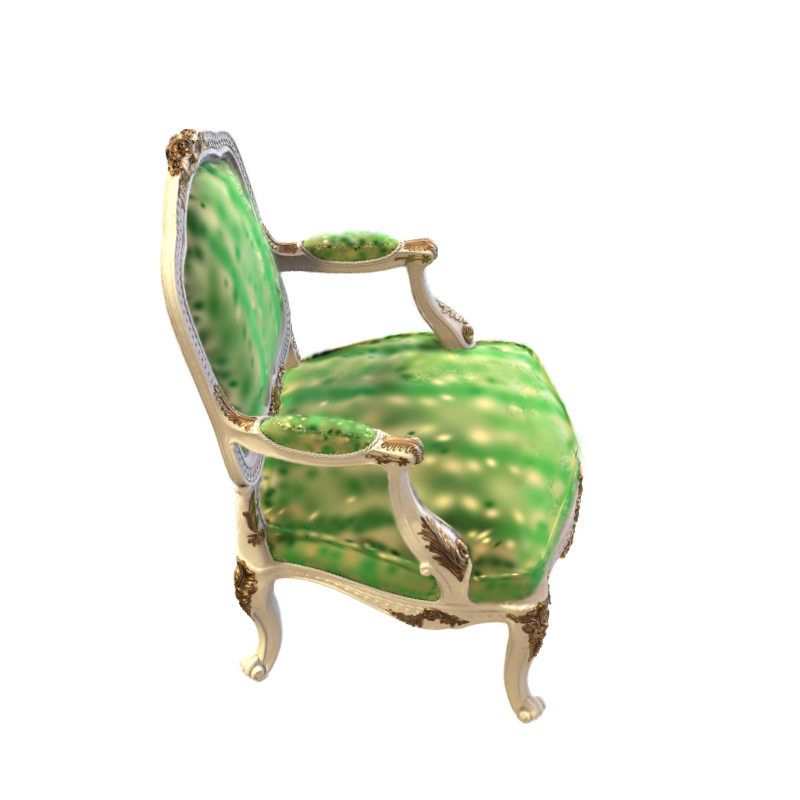} &
\includegraphics[width=0.12\textwidth, trim=150 0 150 120, clip]{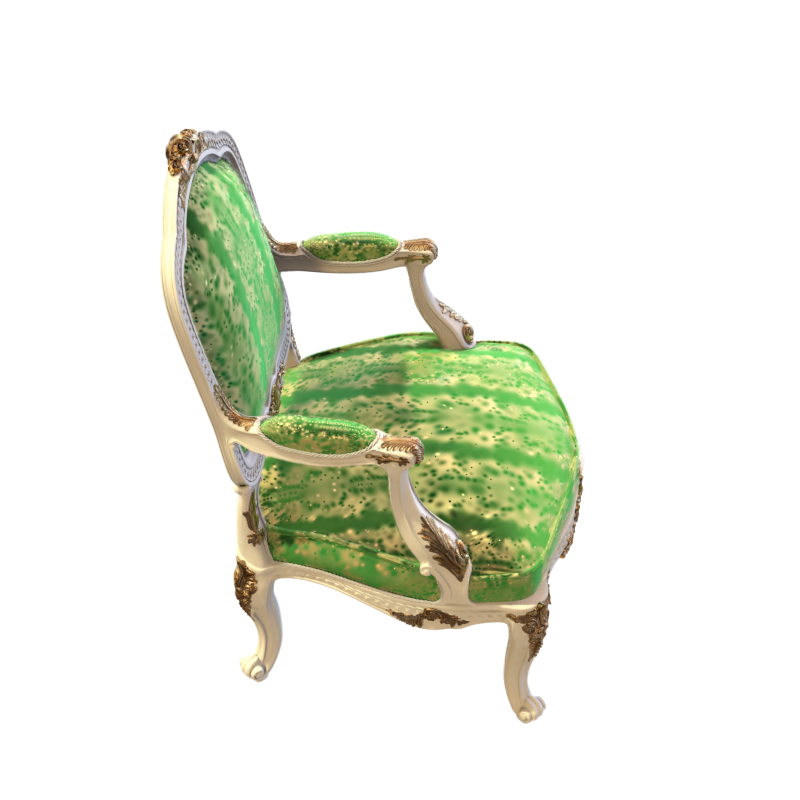}  &
\includegraphics[width=0.12\textwidth, trim=150 0 150 120, clip]{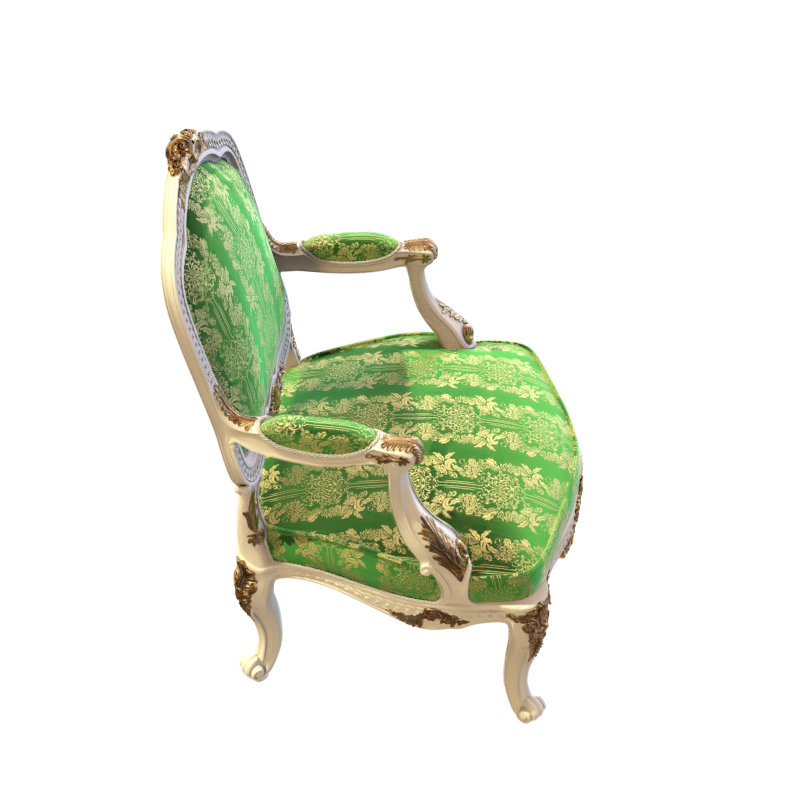} \\
\includegraphics[width=0.12\textwidth, trim=150 0 150 60, clip]{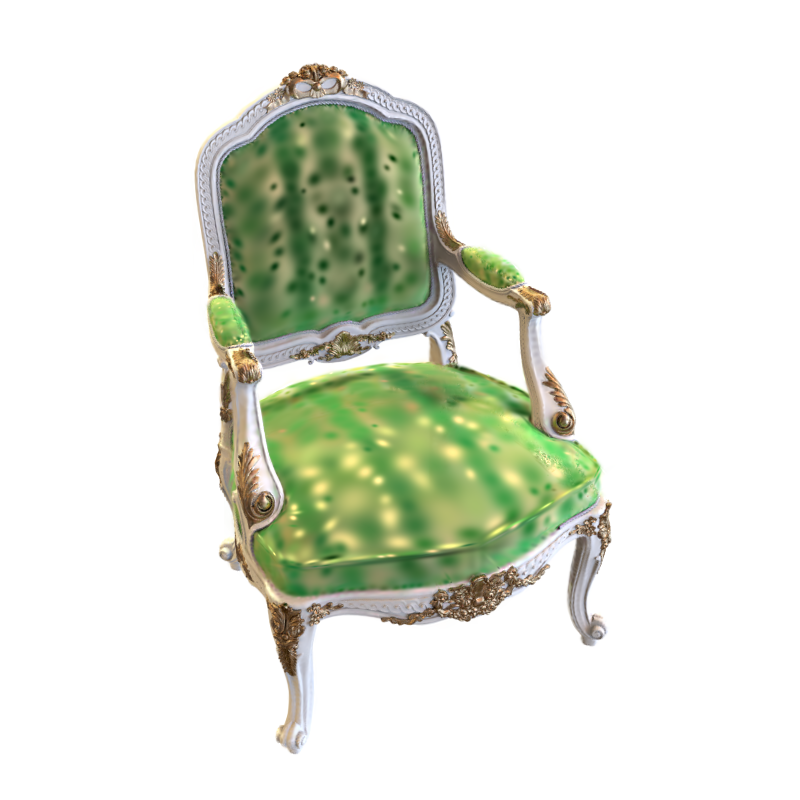} &
\includegraphics[width=0.12\textwidth, trim=150 0 150 60, clip]{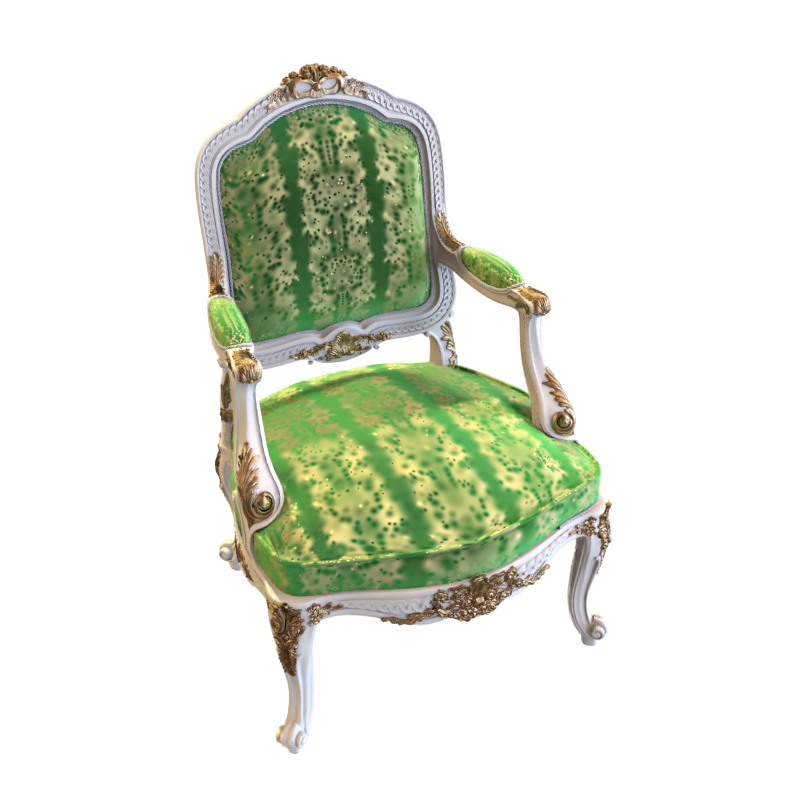}  &
\includegraphics[width=0.12\textwidth, trim=150 0 150 60, clip]{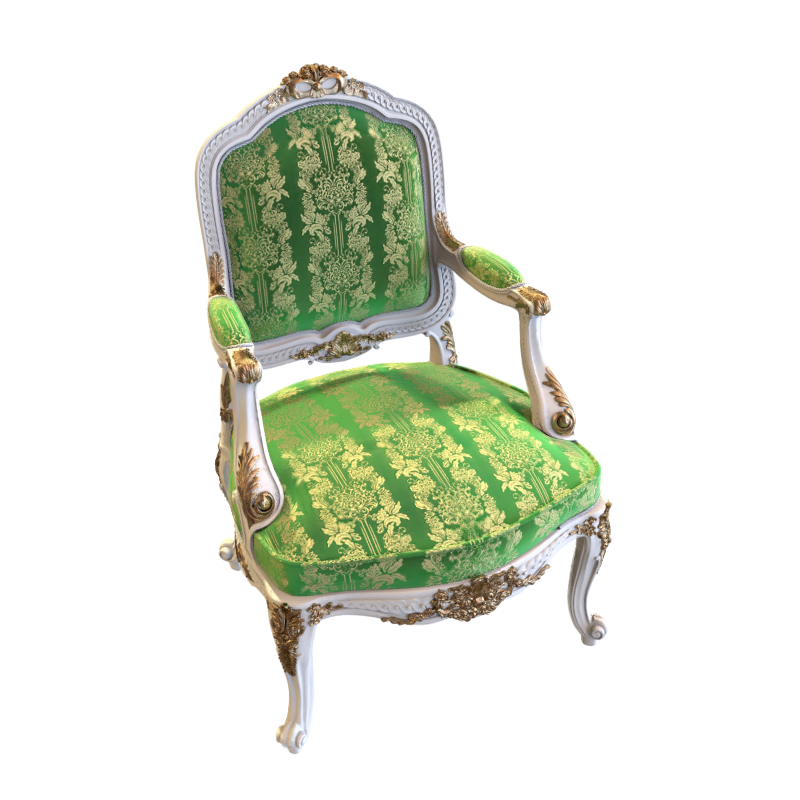} \\
\includegraphics[width=0.12\textwidth, trim=150 0 150 100, clip]{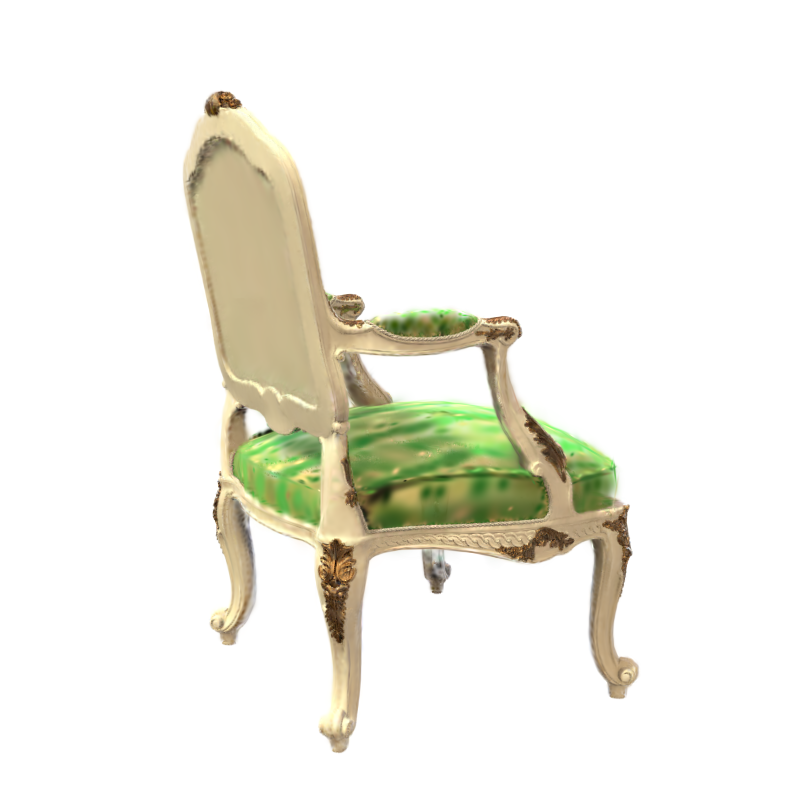} &
\includegraphics[width=0.12\textwidth, trim=150 0 150 100, clip]{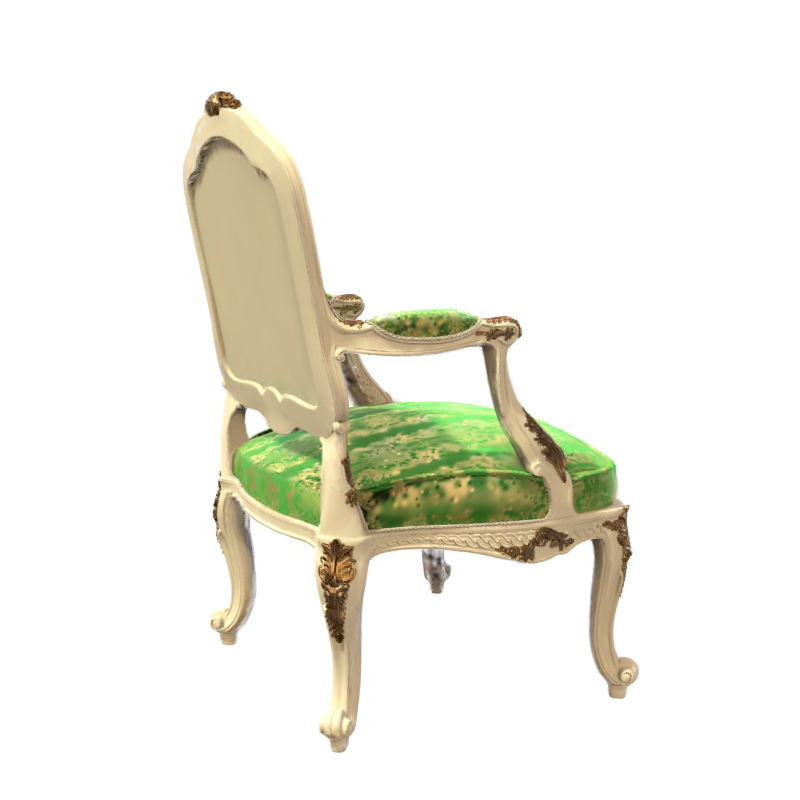}  &
\includegraphics[width=0.12\textwidth, trim=150 0 150 100, clip]{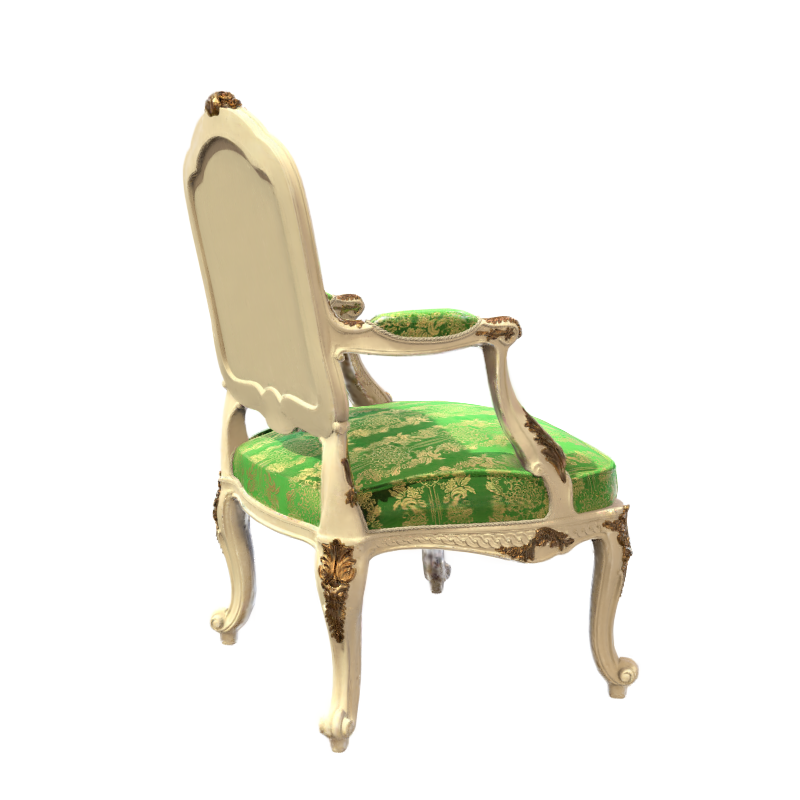} \\
    \end{tabular}
 \caption{A qualitative assessment delineating the impact of varying Gaussian quantities per face, discerned between the initial mesh and its subdivided counterpart.}
\label{fig:chairzaleznoscgaussians} 
\vspace{1cm}
\end{figure}

\begin{figure}[h]
  \centering
  \begin{minipage}[b]{0.48\textwidth}
 \centering
\begin{tikzpicture}
  \node [text width=5cm, align=center,] at (0, 0) {Initial Mesh};
  \node [text width=5cm, align=center,] at (4, 0) {Subdivided Mesh };
\end{tikzpicture}
\begin{tikzpicture}
  \node [text width=4cm, align=center,] at (0, 0) {k=1};
  \node [text width=4cm, align=center,] at (3, 0) {k=10};
  \node [text width=4cm, align=center,] at (5, 0) {k=5};
\end{tikzpicture}

\includegraphics[width=0.95\textwidth, trim = 0 70 0 0, clip]{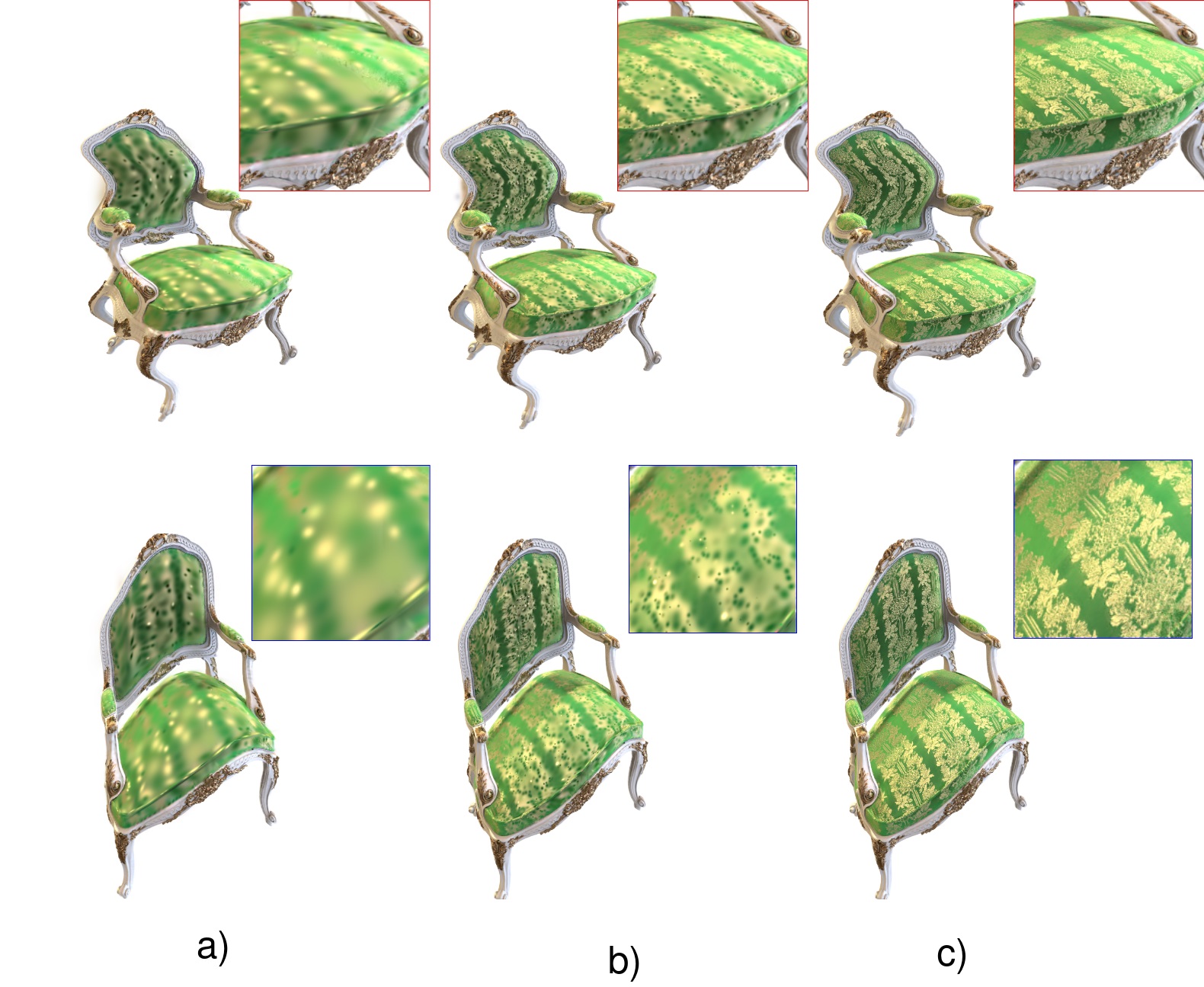}

\caption{A juxtaposition of outcomes is conducted for a model varying in Gaussian counts per face and with a mesh subdivided. \our{} with good mesh initialization is able to capture fine details.}
\label{fig:chairablation} 
  \end{minipage}
  \hfill
  \begin{minipage}[b]{0.48\textwidth}
 	\centering

\quad Before modification \qquad After modification\\
\includegraphics[width=0.45\textwidth, trim=0 0 0 0, clip]{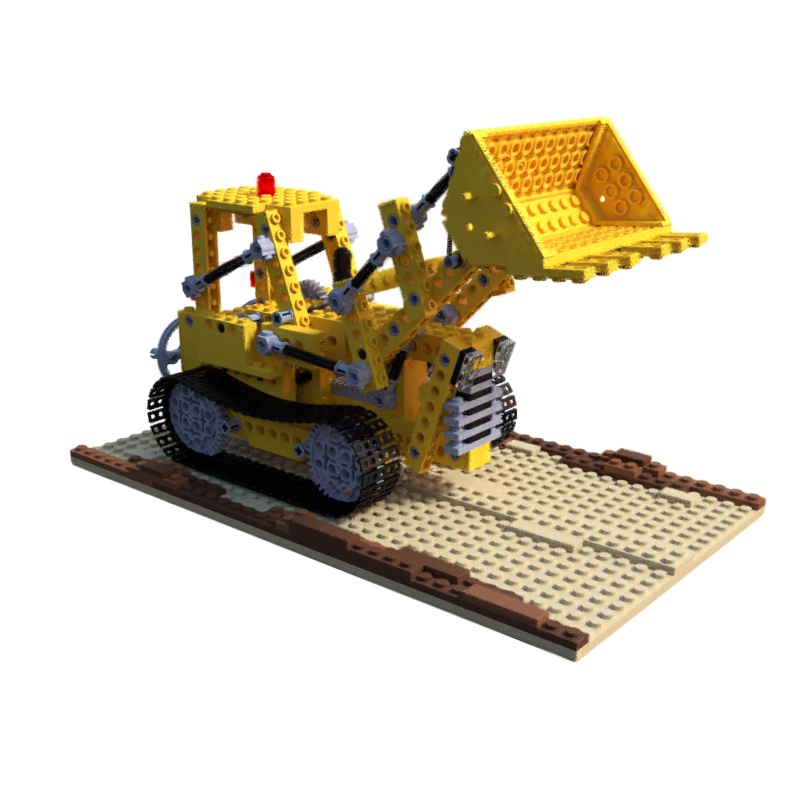} 
\includegraphics[width=0.45\textwidth, trim=0 0 0 0, clip]{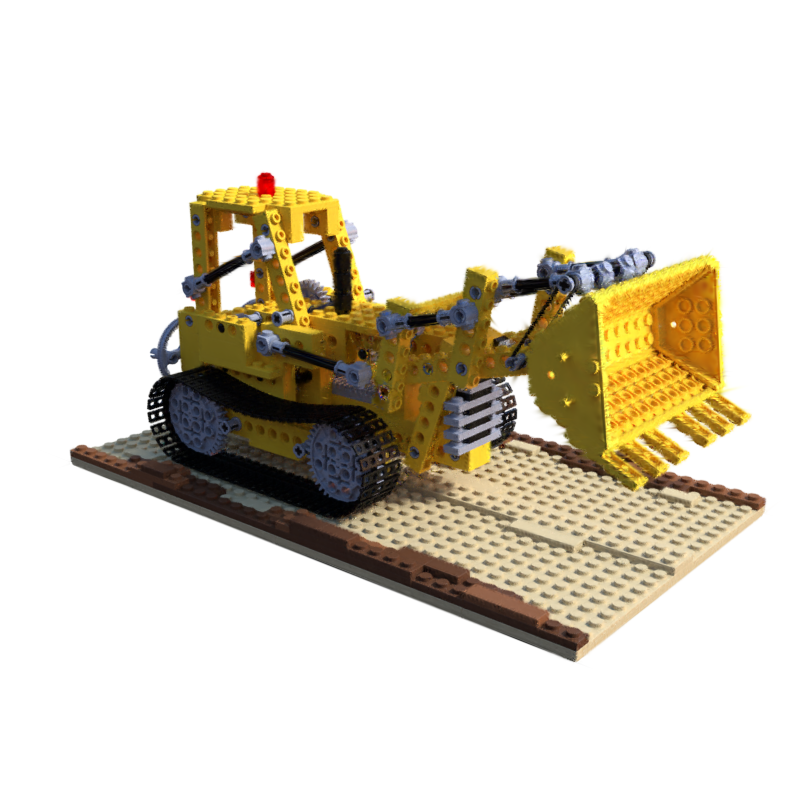}\\
\includegraphics[width=0.45\textwidth, trim=0 0 0 0, clip]{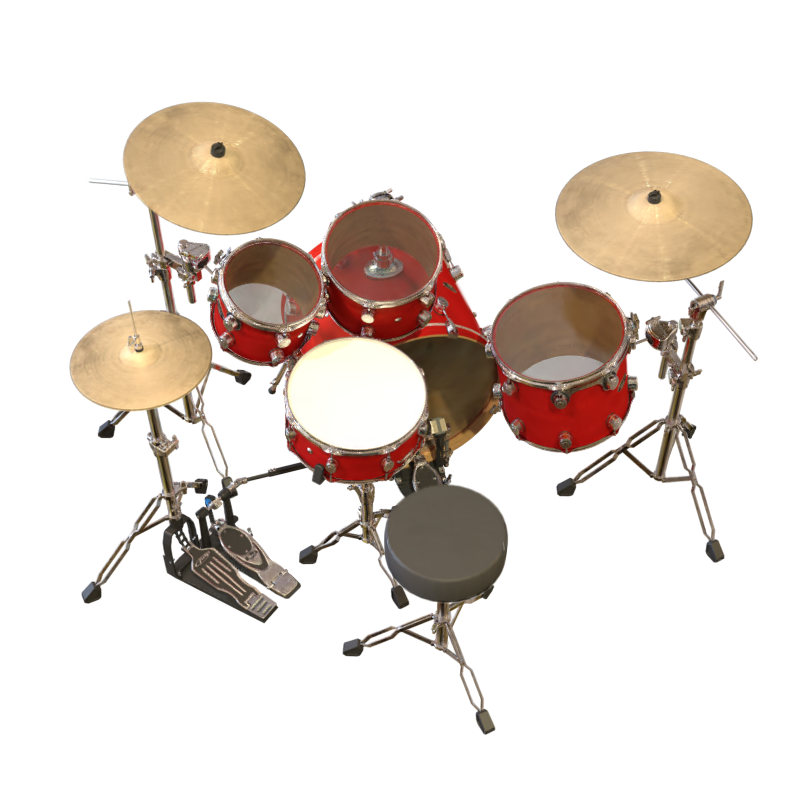} 
\includegraphics[width=0.45\textwidth, trim=0 0 0 0, clip]{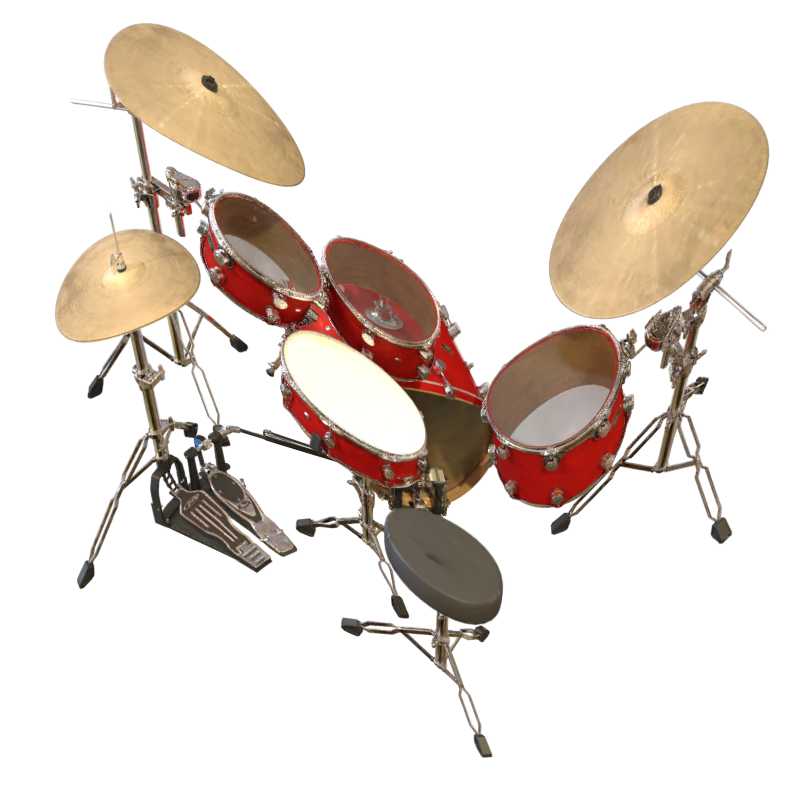}
 \caption{An example of reconstruction and modification using the \our{} model. The model enables more realistic modifications like lifting the excavator blade as well as surrealistic drums bending.}
\label{fig:innemodyfikacje} 
  \end{minipage}
\end{figure}

\subsection{External Experiments}

The appendix to the paper contains experiments related to face dataset and NeRF Synthetic and a continuation of the numerical results expounded in the main document.

\subsection{Mesh fitted}
As shown in Fig. \ref{fig:mesh}, when a well-fitting pre-existing mesh is not available, we can use a FLAME model as the input. With a good fit, adjusting the FLAME parameters allows us to manipulate expressions directly. Since the dataset lacks a well-fitting mesh, we provide a comparative analysis of mesh alignment with NeRFlame (see Fig. \ref{fig:mesh}). Notably, the faces are distinctly delineated and closely resemble the modeled objects, capturing not only postures but also expressions and intricate details. 
Fig. \ref{fig:mesh}

\subsection{Number of Gaussians per face}

\begin{figure}[h!]
 	\centering
    \includegraphics[width=0.45\textwidth]{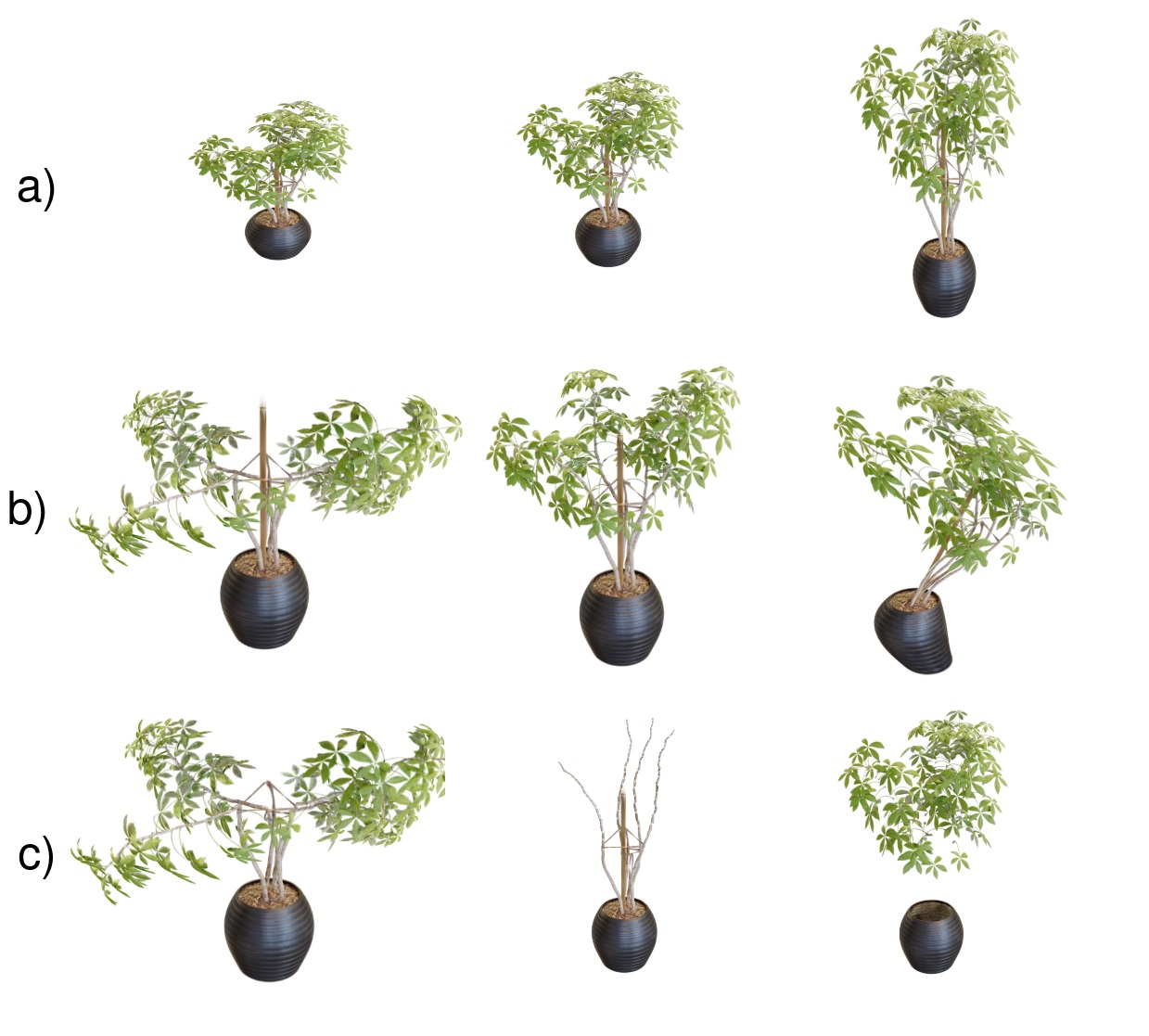}
\caption{Demonstrations of potential modifications facilitated by \our{} include: a) scaling, b) transforming either specific components or entire objects, and c) selectively disregarding specific sections.}
\label{fig:fikusyruchy} 
\vspace{3cm}
\end{figure}

Here, we show the results of the experiments for the NeRF Synthetic dataset with a white background and original mesh. The dependence of the PSNR, SSIM  and LPIPS result on the number of Gaussians per face is shown in Table~\ref{tab:ablationsyntetic}. 

Mostly, as a consequence of increasing the number of Gaussians, we get better results, and we can observe that a single Gaussian per face is insufficient. Materials and Lego has a very dense mesh, consequently positioning an excessive number of Gaussians within a confined space does not yield substantial improvements in the results.

However, experiments have shown that choosing a fixed number of Gaussians per face mesh is inefficient when the mesh contains different face sizes -- in particular, it is difficult to cover large faces. Therefore, we decided to split the large facets. Splitting large facets, even using fewer Gaussians per face, allows for significantly better results and detail capturing. The effects are shown in Fig. \ref{fig:chairzaleznoscgaussians} (initial positions) and Fig. \ref{fig:chairablation} (modifications).

\subsection{Modification}

Illustrating the reconstruction and modification through the application of the \our{} model is shown in Fig.~\ref{fig:innemodyfikacje}. \our{} allows the creation of lifelike alterations, such as elevating the excavator blade or imaginative and surrealistic bends.

In fact, modification depends only on user skills. Fig. \ref{fig:fikusyruchy} shows various modifications of the ficus such as: a) scaling b) modifying a certain part -- like bending branches, or modifying the whole object -- dancing ficus.
(c) ignoring a certain part. All images are generated from the view of the same camera. Note that Gaussians represent color from two sides. Therefore, after removing the soil from the pot, we can still see the inner side of the pot.

\begin{figure}[h!]
 	\centering

\includegraphics[width=0.5\textwidth, trim=0 0 0 0, clip]{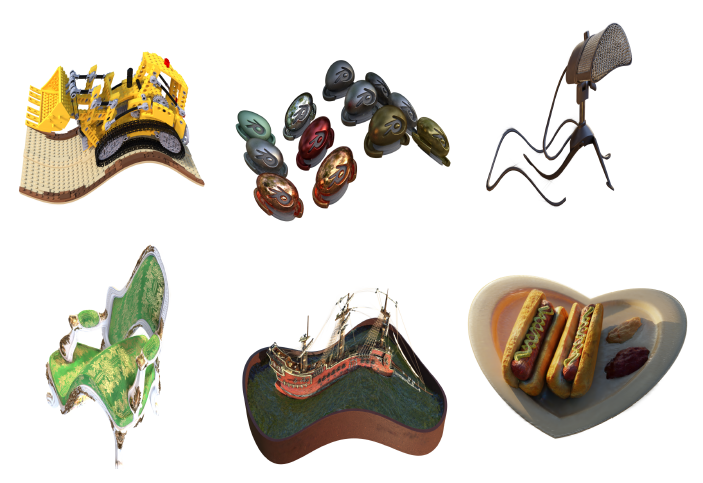}
 \caption{More example of surrealistic modification using the \our{} model}
\label{fig:pseduomeshmod} 
\end{figure}

Fig. \ref{fig:pseduomeshmod} shows that modifications are possible even using only \mesh{}.

\subsection{Extension of numerical results from main paper}

\begin{figure}[h!]
 	\centering
\includegraphics[width=0.5\textwidth=]{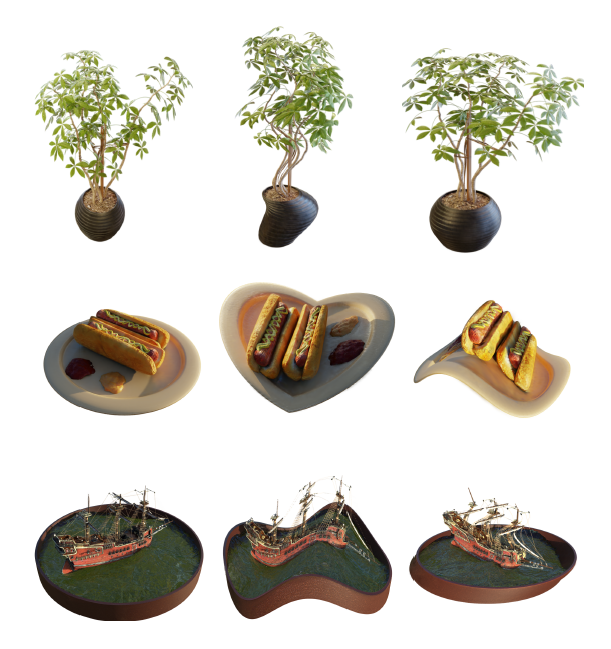} 
 \caption{\our{} produce a hybrid of Gaussian Splatting (GS) and mesh representations. Therefore, \our{} allows real-time modification and adaptation of GS. }
\label{fig:nerfsyntheticslego1} 
\end{figure}

Here, we present extensions of experiments proposed in the main part. In Tab. \ref{tab:ablationsyntetic} the results on NeRF-Synthetic dataset with white background. In this case the \mesh{} approach is dominant across all the metrics. The next table Tab. \ref{tab:mipnerf360} the expanded results on Mip-NeRF360 dataset are presented. This table is the expanded version of the table located in the main paper, where only PSNR values were reported. 
The results of the same experiment conducted on the NeRF-Synthetic dataset with black background can be found in Tab. \ref{tab:nerf}.



\begin{table*}[h]
{\small 
\begin{center}
    \begin{tabular}{@{}l|lllll|llll@{}}
    \multicolumn{10}{c}{PSNR $\uparrow$} \\    
    &  \multicolumn{5}{c}{Outdoor scenes} & \multicolumn{4}{c}{Indoor scenes} \\
    & bicycle & flowers & garden & stump & treehill & room & counter & kitchen & bonsai \\ \hline
    \multicolumn{10}{c}{ Static } \\ \hline
     
Plenoxels   & 21.91 & 20.10 & 23.49 & 20.66 & 22.25 & 27.59 & 23.62 & 23.42 & 24.66 \\
INGP-Big    & 22.17 & 20.65 & 25.07 & 23.47 & 22.37 & 29.69 & 26.69 & 29.48 & 30.69 \\
Mip-NeRF360 & 24.37 & \bf 21.73 & 26.98 & 26.40 & \bf 22.87 & \bf 31.63 & \bf 29.55 & \bf 32.23 & \bf 33.46 \\
GS - 7K & 23.60 & 20.52 & 26.25 & 25.71 & 22.09 & 28.14 & 26.71 & 28.55 & 28.85\\
GS - 30K  & \bf 25.25 & 21.52 & \bf 27.41 & \bf 26.55 & 22.49 & 30.63 & 28.70 & 30.32 & 31.98 \\
\hline
    \multicolumn{10}{c}{ Editable } \\ \hline
R-SuGaR-15K  & 23.14  & - & 25.36 & 24.70 &  - & 30.03  & 27.62 & 29.56 & 30.51\\
\our{} (Our) & \bf 24.99 & \bf 21.27 & \bf 27.22 & \bf 26.54 & \bf 22.39 & \bf 31.52 & \bf 28.92 & \bf 31.12 & \bf 32.09
\\ \hline
    \multicolumn{10}{c}{SSIM $\uparrow$} \\
    &  \multicolumn{5}{c}{Outdoor scenes} & \multicolumn{4}{c}{Indoor scenes} \\
    & bicycle & flowers & garden & stump & treehill & room & counter & kitchen & bonsai \\ \hline
         \multicolumn{10}{c}{ Static } \\ \hline

Plenoxels   & 0.496 & 0.431 & 0.606 & 0.523 & 0.509 & 0.842 & 0.759 & 0.648 & 0.814 \\
INGP-Big    & 0.512 & 0.486 & 0.701 & 0.594 & 0.542 & 0.871 & 0.817 & 0.858 & 0.906 \\
Mip-NeRF360 & 0.685 & 0.583 & 0.813 & 0.744 & 0.632 & 0.913 & 0.894 & 0.920 & \bf 0.941 \\
GS - 7K  &0.675 & 0.525  & 0.836 & 0.728 & 0.598 & 0.884 & 0.873 & 0.900  & 0.910 \\
GS - 30K  & \bf 0.771 & \bf 0.605 & \bf 0.868 & \bf 0.775 & \bf 0.638 & \bf 0.914 & \bf 0.905 & \bf 0.922 & 0.938 \\
\hline
    \multicolumn{10}{c}{ Editable } \\ \hline
R-SuGaR-15K  & 0.640 & - & 0.775 & 0.683  &  - & 0.909  & 0.891 & 0.908 & 0.932\\
\our{} (Our) & \bf 0.757 & \bf 0.596 & \bf 0.860 & \bf 0.766 & \bf 0.628 & \bf 0.916 & \bf 0.904 & \bf 0.923 & \bf 0.939
\\ \hline

    \multicolumn{10}{c}{LPIPS $\downarrow$ }\\ 
    &  \multicolumn{5}{c}{Outdoor scenes} & \multicolumn{4}{c}{Indoor scenes} \\
    & bicycle & flowers & garden & stump & treehill & room & counter & kitchen & bonsai \\ \hline
         \multicolumn{10}{c}{ Static } \\ \hline

Plenoxels   & 0.506 & 0.521 & 0.386 & 0.503 & 0.540 & 0.419 & 0.441 & 0.447 & 0.398 \\
INGP-Big    & 0.446 & 0.441 & 0.257 & 0.421 & 0.450 & 0.261 & 0.306 & 0.195 & 0.205 \\
Mip-NeRF360 & 0.301 & 0.344 & 0.170 & 0.261 & 0.339 & \bf 0.211 & \bf 0.204 & \bf 0.127 & \bf 0.176 \\
GS - 7K &0.318 & 0.417 & 0.153 & 0.287& 0.404 &0.272& 0.254& 0.161& 0.244\\
GS - 30K & \bf 0.205 & \bf 0.336 & \bf 0.103 & \bf 0.210 & \bf 0.317 & 0.220 & \bf 0.204 & 0.129 & 0.205 \\
\hline
    \multicolumn{10}{c}{ Editable } \\ \hline
R-SuGaR-15K  & 0.345 & - & 0.220  & 0.338 &  - & 0.246  & 0.234 & 0.165 & 0.221 \\
\our{} (Our) & \bf 0.216 & \bf 0.342 & \bf 0.111 & \bf 0.219 & \bf 0.332 & \bf 0.221 & \bf 0.203 & \bf 0.129 & \bf 0.205
\\ \hline
    \end{tabular}
    \end{center}
    }
    
    \caption{The quantitative comparisons (PSNR / SSIM / LPIPS) on the Mip-NeRF360 dataset. R-SuGaR-15K, with the number of 1M vertices.}
    \label{tab:mipnerf360}
\end{table*}

\begin{table*}[h]
{\small 
\begin{center}
    \begin{tabular}{@{}l|llllllll@{}}
    \multicolumn{9}{c}{PSNR $\uparrow$} \\
         & Chair & Drums & Lego & Mic & Materials & Ship & Hotdog & Ficus \\ 
 \hline

    \multicolumn{9}{c}{Static} \\ 
 \hline

NeRF & 33.00 & 25.01 & 32.54 & 32.91 & 29.62 & 28.65 & 36.18 & 30.13  \\ 
VolSDF & 30.57 & 20.43 & 29.46 & 30.53 & 29.13 & 25.51 & 35.11 & 22.91 \\ 
Ref-NeRF & 33.98 & 25.43 & 35.10 & 33.65 & 27.10 & 29.24 & 37.04 & 28.74\\ 
ENVIDR & 31.22 & 22.99 & 29.55 & 32.17 & 29.52 & 21.57 & 31.44 & 26.60 \\ 
  Plenoxels 
  & 33.98 & 25.35 & 34.10  & 33.26 & 29.14 &  29.62 & 36.43 & 31.83\\
Gaussian Splatting & \bf 35.82 & \bf 26.17 & \bf 35.69 & \bf 35.34 & \bf 30.00 & \bf 30.87 & \bf 37.67 & \bf 34.83 \\ 
\hline
    \multicolumn{9}{c}{Editable} \\ 
 \hline
RIP-NeRF     &34.84 & 24.89 & 33.41 & 34.19 &28.31 & 30.65 & 35.96 & 32.23\\
\our{} (our) &  \bf 35.38 & \bf 26.03 & \bf 35.89 & \bf 37.16 & 29.62 & \bf 31.55 & 37.56 & 35.12\\
\our{} (our - \mesh{}) & 34.74 & 25.94 & 35.88 & 36.58 & \bf 30.50 & 30.66 & \bf 37.84 & \bf 35.41 \\
  \vspace{0.5cm}\\
    \multicolumn{9}{c}{SSIM $\uparrow$} \\
     \hline

    \multicolumn{9}{c}{Static} \\ 
 \hline
NeRF & 0.967 & 0.925 & 0.961 & 0.980 & 0.949 & 0.856 & 0.974 & 0.964  \\
VolSDF  & 0.949 & 0.893 & 0.951 & 0.969 & 0.954 & 0.842 & 0.972 & 0.929  \\
Ref-NeRF & 0.974 & 0.929 & 0.975 & 0.983 & 0.921 & 0.864 & 0.979 & 0.954 \\
ENVIDR & 0.976 & 0.930 & 0.961 & 0.984 &  \bf 0.968 & 0.855 & 0.963 & \bf 0.987 \\ 
Plenoxels & 0.977 & 0.933 & 0.975 & 0.985 & 0.949 & 0.890 & 0.980 & 0.976 \\
Gaussian Splatting &  \bf 0.987 & \bf 0.954 & \bf 0.983 & \bf 0.991 & 0.960 & \bf 0.907 &  \bf 0.985 & \bf 0.987 \\ 
\hline
    \multicolumn{9}{c}{Editable} \\ 
 \hline
RIP-NeRF     & 0.980& 0.929 & 0.977 & 0.962 & 0.943 & \bf 0.916 & 0.963 & 0.979 \\
\our{} (our)  & \bf 0.987 & \bf 0.953 & \bf 0.982 & \bf 0.992 & 0.952 & 0.904 & \bf 0.985 & \bf 0.986\\
\our{} (our - \mesh{}) & 0.983 & 0.945 & 0.981 & \bf 0.992 & \bf 0.958 & 0.888 & \bf 0.985 & \bf 0.986 \\
\vspace{0.5cm} \\
     \multicolumn{9}{c}{LPIPS $\downarrow$ }\\ 
NeRF & 0.046 & 0.091 & 0.050 & 0.028 & 0.063 & 0.206 & 0.121 & 0.044 \\
VolSDF & 0.056 & 0.119 & 0.054 & 0.191 & 0.048 & 0.191 & 0.043 & 0.068\\
Ref-NeRF & 0.029 & 0.073 & 0.025 & 0.018 & 0.078 & 0.158 & 0.028 & 0.056 \\
Plenoxels & 0.031 & 0.067 & 0.028 &  0.015 & 0.057 & 0.134 & 0.037 & 0.026 \\
ENVIDR & 0.031 & 0.080 & 0.054 & 0.021 & 0.045 & 0.228 & 0.072 &  \bf 0.010  \\ 
Gaussian Splatting &  \bf 0.012 & \bf 0.037 & \bf 0.016 & \bf 0.006 & \bf 0.034 & \bf 0.106 & \bf 0.020 & 0.012 \\
\hline
    \multicolumn{9}{c}{Editable} \\ 
 \hline
RIP-NeRF     & - & - & -& - & -& - & - & -\\
\our{} (our)  & \bf 0.009 & \bf 0.038 & \bf 0.014 & \bf 0.005 & 0.042 & \bf 0.090 & \bf 0.017 & \bf 0.012\\
\our{} (our - \mesh{}) & 0.012 & 0.039 & 0.016 & 0.006 & \bf 0.036 & 0.109 & 0.018  & \bf 0.012 \\
    \end{tabular}
    \end{center}
    }
    
    \caption{Quantitative comparisons (PSNR/SSIM/LPIPS) on a NeRF-Synthetic dataset showing that \our{} gives comparable results with other models.}
    \label{tab:nerf}
\end{table*} 

\end{document}